\pdfoutput=1

\PassOptionsToPackage{table}{xcolor}  
\documentclass[11pt]{article}

\usepackage[final]{acl}

\usepackage{times}
\usepackage{latexsym}

\usepackage[T1]{fontenc}

\usepackage[utf8]{inputenc}

\usepackage{microtype}

\usepackage{inconsolata}

\usepackage{graphicx}

\usepackage{booktabs}
\usepackage{multirow}
\usepackage{graphicx}
\usepackage{colortbl}

\usepackage{lipsum} 
\usepackage{booktabs}
\usepackage{multirow}
\usepackage{graphicx}
\usepackage[normalem]{ulem}
\useunder{\uline}{\ul}{}
\usepackage{amsmath}
\usepackage{amssymb}
\usepackage{bbm}
\usepackage{afterpage}
\usepackage{xspace}

\usepackage{pifont}

\newcommand{\MethodName}{GLEAN\xspace}

\title{GLEAN: Active Generalized Category Discovery with Diverse LLM Feedback}

\author{
 \textbf{Henry Peng Zou\textsuperscript{1}\thanks{Work done as an intern at Amazon.}},
 \textbf{Siffi Singh\textsuperscript{2}},
 \textbf{Yi Nian\textsuperscript{2}},
 \textbf{Jianfeng He\textsuperscript{2}},
\\
 \textbf{Jason Cai\textsuperscript{2}},
 \textbf{Saab Mansour\textsuperscript{2}},
 \textbf{Hang Su\textsuperscript{2}},
\\
 \textsuperscript{1}University of Illinois Chicago,
 \textsuperscript{2}AWS AI Labs
\\
 \texttt{
   pzou3@uic.edu
 }
}

\begin{document}
\maketitle
\begin{abstract}
\textit{Generalized Category Discovery} (GCD) is a practical and challenging open-world task that aims to recognize both known and novel categories in unlabeled data using limited labeled data from known categories. Due to the lack of supervision, previous GCD methods face significant challenges, such as difficulty in rectifying errors for confusing instances, and inability to effectively uncover and leverage the semantic meanings of discovered clusters. Therefore, additional annotations are usually required for real-world applicability. However, human annotation is extremely costly and inefficient. To address these issues, we propose \MethodName
, a unified framework for generalized category discovery that actively learns from diverse and collaborative LLM feedback. Our approach leverages three different types of LLM feedback to: (1) improve instance-level contrastive features, (2) generate category descriptions, and (3) align uncertain instances with LLM-selected category descriptions. Extensive experiments demonstrate the superior performance of \MethodName over state-of-the-art models across diverse datasets, metrics, and supervision settings. 
Code is available at \href{https://github.com/amazon-science/Glean}{GitHub}.

\end{abstract}

\section{Introduction}

The success of many deep learning models often heavily depends on some ideal assumptions, such as the availability of large amounts of labeled data, and the closed-world setting where unlabeled data shares the same set of pre-defined categories as labeled data \cite{Zhong_2021_CVPR, zhang-etal-2022-new, an2023generalized, zou-caragea-2023-jointmatch}. However, these assumptions often do not hold in many real-world scenarios. For example, in customer service intent detection, new types of inquiries may emerge over time \cite{tang-etal-2023-rsvp, zhang-etal-2024-discrimination}. Similarly, e-commerce product categorization faces ongoing introduction of novel product types \cite{gong-etal-2023-transferable, zou-etal-2024-implicitave, zou-etal-2024-eiven}. In this work, we try to lift these assumptions by considering a more realistic and challenging setting: \textit{Generalized Category Discovery} (GCD) \cite{vaze2022generalized}.

\begin{figure}[!t]
    \centering
    \includegraphics[width=\columnwidth]{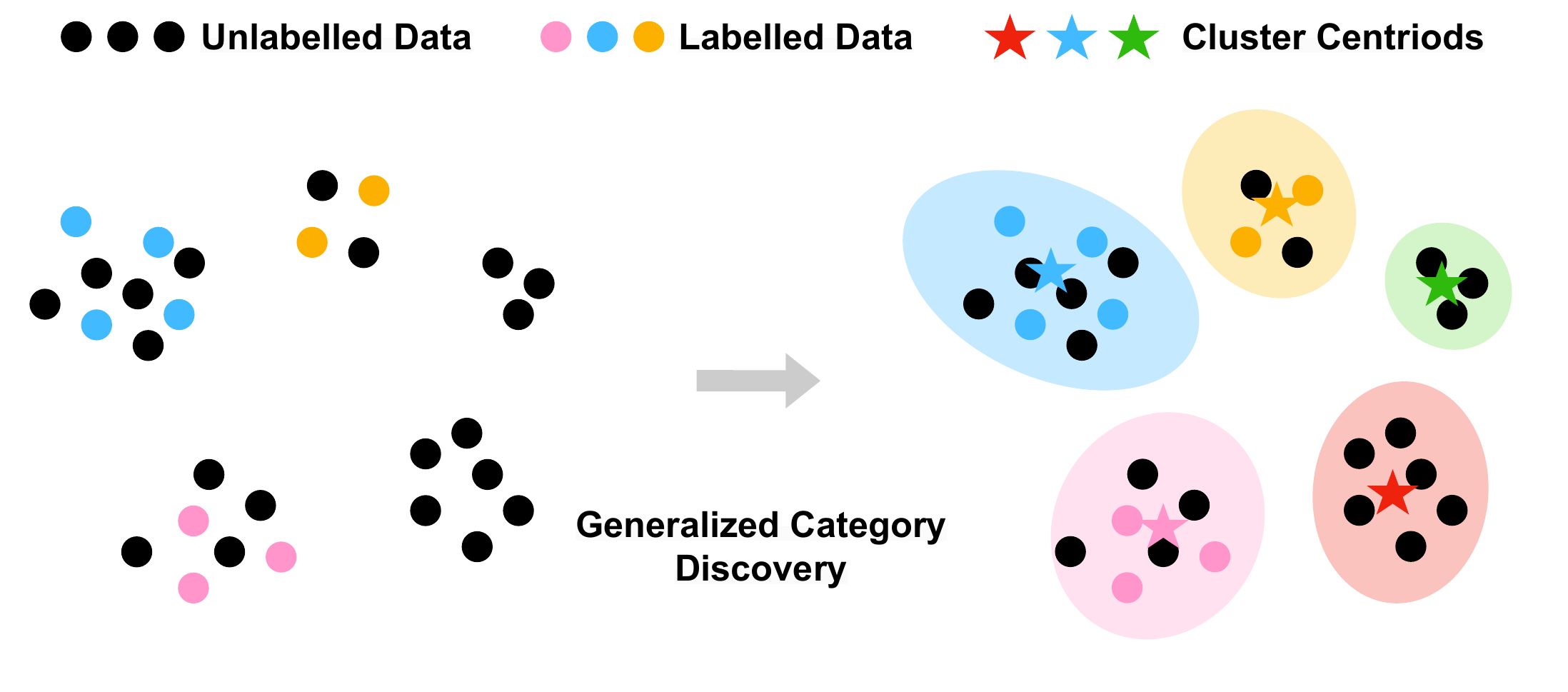} 
    \caption{\textit{Generalized Category Discovery} aims to automatically categorize unlabeled data by leveraging the information from a limited number of labeled data from known categories, while the unlabeled data may come from both known and novel categories.}
    \label{fig:gcd}
\end{figure}

As illustrated in Figure \ref{fig:gcd}, GCD addresses a scenario where only a portion of the dataset is labeled and only a subset of categories is known. It aims to automatically categorize all unlabeled data, including instances from both known and novel categories, by leveraging information from a limited number of labeled instances \cite{vaze2022generalized, wen2023simgcd}. This task is particularly relevant in realistic dynamic environments where new categories emerge over time, and manually labeling all data is impractical or prohibitively expensive \cite{ma2024active, zhang-etal-2023-clusterllm}.

Previous work on GCD \cite{vaze2022generalized, pu2023dynamic} has primarily focused on applying contrastive learning to both labeled and unlabeled data to learn discriminative representations, followed by clustering methods like K-Means++ to discover both seen and novel categories. However, these approaches face inherent challenges: (1) Due to the lack of supervision, their models struggle to correct errors for confusing instances and categories. (2) Moreover, these methods often fail to uncover and leverage the semantic meanings of discovered clusters effectively.

To address the aforementioned issues and limitations, we propose \MethodName, a unified framework for generalized category discovery that actively learns from diverse and quality-enhanced LLM feedback. Our approach leverages LLMs in three effective ways: (1) Similar Instance Selection: We use LLMs to identify similar instances among ambiguous data points, refining embeddings through neighborhood contrastive learning \cite{zhong2021neighborhood}. (2) Category Characterization: LLMs generate interpretable names and descriptions for newly discovered categories, making novel categories more accessible and meaningful. (3) Pseudo Category Selection and Alignment: We associate instance embeddings with LLM-selected category descriptions, fostering improved representation learning that considers category-instance relationships. By incorporating these diverse forms of LLM feedback, \MethodName achieves significant performance improvements over existing methods across multiple benchmark datasets. We also analyze the performance of \MethodName with varying numbers of known categories. Detailed ablation studies and analyses are provided to further understand each component and hyperparameter.

\begin{figure*}[!tbh]
    \centering
    \includegraphics[width=\textwidth]{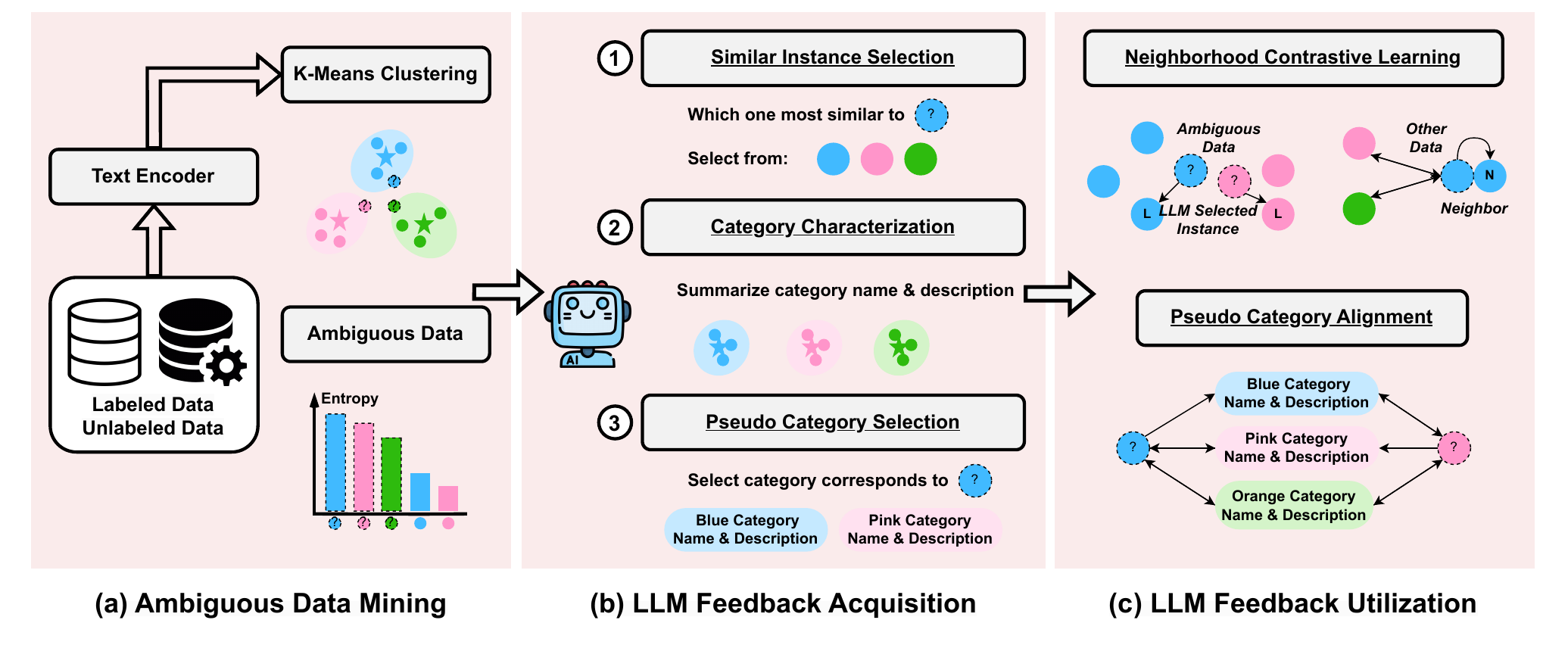}
    \caption{Pipeline of \MethodName. Both labeled and unlabeled data are first forwarded to a text encoder to extract features for K-Means++ clustering. Then we compute entropy and select instances with high entropy as ambiguous data to obtain LLM feedback for further refinement. Specifically, we query LLM to (1) select similar instances, (2) generate category descriptions and (3) assign pseudo categories to ambiguous data. Lastly, the three diverse feedback types are leveraged for model training via neighborhood contrastive learning and pseudo category alignment. During inference, we only utilize the text encoder and obtain final results via K-Means++ clustering on the extracted features. Illustration of the three types of LLM feedback with concrete examples is provided in Figure \ref{fig:llm_feedback}.}
    \label{fig:pipeline_illustration}
\end{figure*}

\section{Related Work}

\noindent \textbf{Generalized Category Discovery. }
Generalized Category Discovery (GCD) \cite{vaze2022generalized, wen2023simgcd, an-etal-2024-generalized, liang-etal-2024-actively} is a recently emerged task that addresses a realistic scenario where only limited labeled data is available and new categories may emerge. The goal is to automatically cluster all unlabeled data from both seen and novel categories \cite{bai2023towards, zou-etal-2023-decrisismb}. Pioneering works \cite{vaze2022generalized, pu2023dynamic} employ supervised and unsupervised contrastive learning to obtain discriminative embeddings, followed by clustering methods such as K-Means++. \citet{wen2023simgcd, bai2023towards} propose leveraging parametric classifiers and soft pseudo labels to mitigate model bias towards seen categories, thereby enhancing overall performance. Despite these advancements, traditional GCD methods that do not actively leverage additional human or LLM supervision signals \cite{zhang2021discovering, zhang-etal-2022-new, zhou-etal-2023-probabilistic, zhang2023clustering, liang-liao-2023-clusterprompt, Raedt2023IDASID, sung-etal-2023-pre} still face significant challenges, including difficulty in rectifying errors for confusing instances and an inability to leverage the semantic meanings of discovered clusters \cite{ma2024active, an-etal-2024-generalized}. \\

\noindent \textbf{Active Learning and GCD. }
Active Learning (AL) \cite{ren2021survey, Ma2024ActiveGC} aims to improve model performance by selecting and annotating a limited number of informative samples. Diverse sample selection strategies have been proposed, including uncertainty-based methods \cite{wang2014new, zhang-etal-2023-clusterllm}, diversity-based methods \cite{sener2017active, Ash2020Deep} and hybrid methods \cite{agarwal2020contextual, huang2010active}. However, annotation in traditional AL is usually expensive and time-consuming \cite{cheng-etal-2023-improving, zhang-etal-2023-clusterllm}. Recent work in GCD, such as ALUP \cite{an-etal-2024-generalized} and Loop \cite{liang-etal-2024-actively}, has started to leverage LLMs feedback as a cost-effective alternative to human annotators to provide additional supervision signal for ambiguous data. However, these approaches utilize only a single type of LLM feedback in their model learning processes or do not account for feedback quality \cite{an-etal-2024-generalized, liang-etal-2024-actively, liang-etal-2024-synergizing}. These limitations restrict their ability to effectively refine model predictions, particularly in ambiguous or challenging cases. To address these issues, we propose a unified framework for GCD that actively learns from diverse and quality-enhanced LLM feedback. To the best of our knowledge, this is the first work to consider both the diversity and quality of LLM feedback in the context of GCD. Table \ref{tab:comp_closely_related_work} provides a detailed comparison of our approach and closely related work in terms of LLM feedback diversity and quality.


\section{\MethodName}

\subsection{Problem Formulation and Overview}
\noindent \textbf{Problem Formulation. }
Formally, assume we have an open-world dataset $\mathcal{D}$, comprising two subsets: a \underline{l}abeled set $\mathcal{D}_l=\{ (x_i, y_i)\}_{i=1}^{N_l}$ contains only known categories, and an \underline{u}nlabeled set $\mathcal{D}_u=\{x_i\}_{i=1}^{N_u} $ contains both known and novel categories. \textit{Generalized Category Discovery} (GCD) aims to accurately categorize all unlabeled data in $D_u$, having access to labels in $\mathcal{D}_l$. The total number of categories $K$ is regarded as a known prior \cite{wen2023simgcd, an-etal-2024-generalized}.\\

\noindent \textbf{Overview. }
The main pipeline for \MethodName is shown in Figure \ref{fig:pipeline_illustration}. Both labeled data $\mathcal{D}_l=\{ (x_i, y_i)\}_{i=1}^{N_l}$ and unlabeled data $\mathcal{D}_u=\{x_i\}_{i=1}^{N_u}$ are first forwarded to a backbone encoder $f(\cdot)$ to extract initial features $h_i = f(x_i)$. These features are then passed through a projection head $g(\cdot)$ to learn contrastive features $z_i = g(h_i)$. We mine ambiguous data and leverage diverse LLM feedback to: (1) refine their contrastive features, (2) generate category descriptions, and (3) align ambiguous instances with LLM-selected category descriptions. During inference, we only utilize the backbone and obtain final results via K-Means++ clustering on the post-backbone features. Next, we explain each main component and how we acquire and utilize each type of LLM feedback in detail.

\begin{figure}[!t]
    \centering
    \includegraphics[width=\columnwidth]{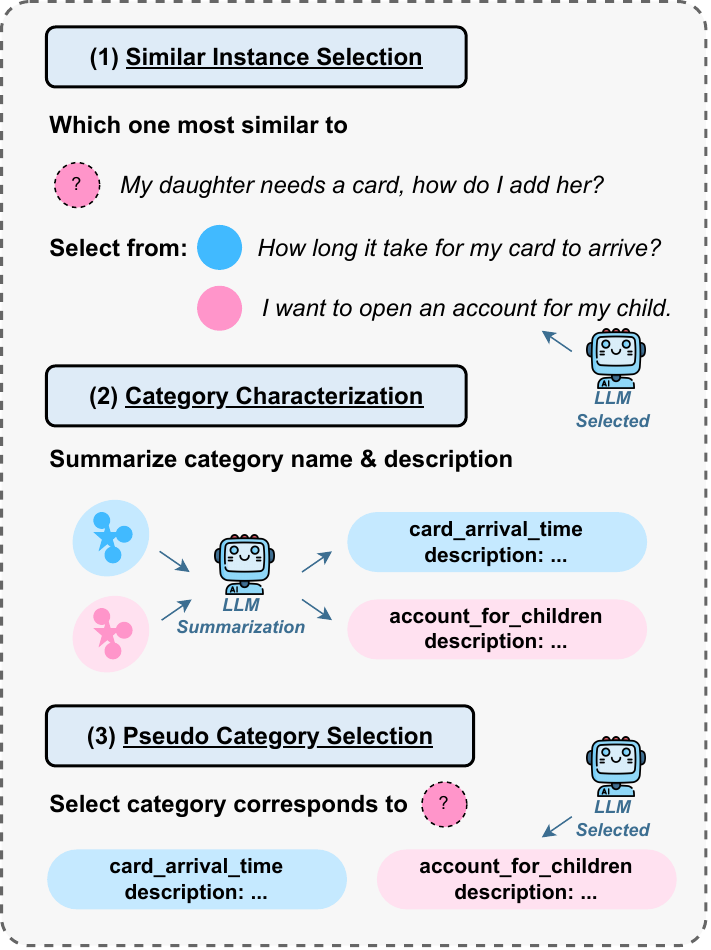} 
    \caption{Illustration of three different types of LLM feedback utilized in \MethodName. Illustration of the whole pipeline is provided in Figure \ref{fig:pipeline_illustration}.}
    \label{fig:llm_feedback}
\end{figure}

\subsection{LLM Feedback 1: Similar Instance Selection and Utilization}
\label{sec:feedback1}


Following \cite{liang-etal-2024-actively, chen2025semi}, we mine ambiguous data with certainty consideration and potential positive candidates to query LLM for similar instance selection and refine their embeddings to be more contrastive. \\

\label{sec:ambiguous_data_mining}
\noindent \textbf{Ambiguous Data Mining. }
We mine ambiguous instances based on entropy. Specifically, we first extract features from the backbone and perform clustering via K-Means++. The soft assignment of each sample $x_i$ to each cluster $k$ is calculated with the commonly used Student's $t$-distribution \cite{xie2016unsupervised} as: 
\begin{equation}
    p_{ik} = \frac{(1 + \|h_i - \mu_k\|^2/\alpha)^{-\frac{\alpha+1}{2}}}{\sum_{k'} (1 + \|h_i - \mu_{k'}\|^2/\alpha)^{-\frac{\alpha+1}{2}}}
\end{equation}
where $h_i$ is the extracted feature, $\mu_k$ is the cluster center from K-Means++, $\alpha$ is the degree of freedom in the Student's $t$-distribution. We then compute entropy to measure the uncertainty for each sample: 
\begin{equation}
    \mathcal{H}_i = - \sum_{k=1}^{K} p_{ik} \log p_{ik}
\end{equation}
The data with the highest entropy is regarded as ambiguous data and is used to form the query set:
\begin{equation}
    \mathcal{Q} = \{ x_i | \mathcal{H}_i \in top_v(\mathcal{H}) \}
\end{equation}
where $v$ is a hyperparameter that determines the total amount of samples to query LLMs.\\

\label{sec:similar_instance_selection}
\noindent \textbf{Similar Instance Selection. }
For each ambiguous data, we randomly sample $M$ instances from each of its closest $M$ clusters as candidates. We then query the LLM to choose the instance most similar to the ambiguous data as the \textit{positive} instance for the following neighborhood contrastive learning. The corresponding prompt and examples are provided in Figure \ref{fig:feedback1_prompt_example}. For data other than ambiguous data, we do not query the LLM but randomly sample an instance from its $k$-nearest neighbors and regard it as the \textit{positive} instance.  \\

\noindent  \textbf{Neighborhood Contrastive Learning. } 
Lastly, we finetune our model to refine embeddings to be more contrastive via the neighborhood contrastive learning loss \cite{zhong2021neighborhood}:

\begin{equation}
\mathcal{L}^{\mathrm{ncl}} = -\frac{1}{N} \sum_{i=1}^N \log \frac{e^{ \left(z_i^T z_{p_i} / \tau\right)}}{\sum_{j \neq i} e^{ \left(z_i^T z_j / \tau\right)}}
\end{equation}
where $p_i$ is the LLM-selected \textit{positive} instance for the mined ambiguous data or the sampled positive neighbor for other data, $z_j$ is the embedding of other in-batch data, $\tau$ is the temperature parameter.

\subsection{LLM Feedback 2: Category Characterization}
\label{sec:feedback2}

Upon identifying the ambiguous data in the previous step, we can query the LLM to obtain pseudo category labels for these samples and use them to enhance model performance. However, since the category labels of newly emerged categories remain unknown, it is infeasible to request LLMs to directly assign the selected data to unknown categories. Thus we propose \textit{\textbf{Category Characterization}} to characterize the novel categories and enable LLM to assign pseudo category labels. Specifically: (1) After obtaining the K-Means++ clustering results on post-backbone features, we first select top-$k$ samples closest to each cluster centroid as representative data. (2) Then for each cluster, we use its representative data to query LLM to generate a pair of cluster name and description. The corresponding prompt and example are provided in Figure \ref{fig:feedback2_prompt_example}. Comprehensive analyses are conducted to measure the performance of category characterization and the effect of different representative data selection strategies, which are presented in Appendix \ref{appendix:category_characterization}. Notably, using only 10 representatives, we achieve 77\% coverage and 71\% semantic matching score.

\subsection{LLM Feedback 3: Pseudo Category Selection and Alignment}
\label{sec:feedback3}

Previous LLM-assisted GCD approaches \cite{liang-etal-2024-actively, an-etal-2024-generalized} only leverage instance-level relationship for model optimization. Here we take category-instance relationship into account and leverage category-instance LLM feedback to align ambiguous instances with LLM-selected positive category names \& descriptions.

\begin{table*}[!tbh]
\centering
\resizebox{\textwidth}{!}{%
\begin{tabular}{clcccccccccc}
\toprule
\multicolumn{1}{l}{} &  & \multicolumn{3}{c}{\textbf{CLINC}} & \multicolumn{3}{c}{\textbf{BANKING}} & \multicolumn{3}{c}{\textbf{StackOverflow}} & \multicolumn{1}{l}{} \\ \midrule
\multicolumn{1}{c|}{\textbf{KCR}} & \multicolumn{1}{l|}{\textbf{Methods}} & \textbf{ACC} & \textbf{ARI} & \multicolumn{1}{c|}{\textbf{NMI}} & \textbf{ACC} & \textbf{ARI} & \multicolumn{1}{c|}{\textbf{NMI}} & \textbf{ACC} & \textbf{ARI} & \multicolumn{1}{c|}{\textbf{NMI}} & \textbf{Average} \\ \midrule
\multicolumn{1}{c|}{} & \multicolumn{1}{l|}{GCD (CVPR 2022)} & 83.29 & 76.77 & \multicolumn{1}{c|}{93.22} & 21.17 & 9.35 & \multicolumn{1}{c|}{43.41} & 17.00 & 3.42 & \multicolumn{1}{c|}{14.57} & 40.24 \\
\multicolumn{1}{c|}{} & \multicolumn{1}{l|}{SimGCD (ICCV 2023)} & 83.24 & 75.89 & \multicolumn{1}{c|}{92.79} & 25.62 & 12.67 & \multicolumn{1}{c|}{47.46} & 18.50 & 6.49 & \multicolumn{1}{c|}{17.91} & 42.29 \\
\multicolumn{1}{c|}{} & \multicolumn{1}{l|}{Loop (ACL 2024)} & 84.89 & 77.43 & \multicolumn{1}{c|}{93.26} & 21.56 & 10.24 & \multicolumn{1}{c|}{44.77} & 18.80 & 5.76 & \multicolumn{1}{c|}{17.54} & 41.58 \\
\multicolumn{1}{c|}{\multirow{-4}{*}{5\%}} & \multicolumn{1}{l|}{\cellcolor{blue!18}\textbf{\MethodName (Ours)}} & \cellcolor{blue!18}\textbf{88.18} & \cellcolor{blue!18}\textbf{82.40} & \multicolumn{1}{c|}{\cellcolor{blue!18}\textbf{94.94}} & \cellcolor{blue!18}\textbf{30.94} & \cellcolor{blue!18}\textbf{18.32} & \multicolumn{1}{c|}{\cellcolor{blue!18}\textbf{54.05}} & \cellcolor{blue!18}\textbf{22.30} & \cellcolor{blue!18}\textbf{8.32} & \multicolumn{1}{c|}{\cellcolor{blue!18}\textbf{21.25}} & \cellcolor{blue!18}\textbf{46.74} \\ \midrule
\multicolumn{1}{c|}{} & \multicolumn{1}{l|}{GCD (CVPR 2022)} & 82.04 & 75.95 & \multicolumn{1}{c|}{93.33} & 59.09 & 46.34 & \multicolumn{1}{c|}{76.22} & 75.40 & 56.01 & \multicolumn{1}{c|}{72.66} & 70.78 \\
\multicolumn{1}{c|}{} & \multicolumn{1}{l|}{SimGCD (ICCV 2023)} & 84.71 & 77.08 & \multicolumn{1}{c|}{93.27} & 60.03 & 47.80 & \multicolumn{1}{c|}{76.53} & 77.10 & 57.70 & \multicolumn{1}{c|}{72.30} & 71.84 \\
\multicolumn{1}{c|}{} & \multicolumn{1}{l|}{Loop (ACL 2024)} & 84.89 & 78.12 & \multicolumn{1}{c|}{93.52} & 64.97 & 53.05 & \multicolumn{1}{c|}{79.14} & 80.50 & \textbf{62.97} & \multicolumn{1}{c|}{75.98} & 74.79 \\
\multicolumn{1}{c|}{\multirow{-4}{*}{10\%}} & \multicolumn{1}{l|}{\cellcolor{blue!18}\textbf{\MethodName (Ours)}} & \cellcolor{blue!18}\textbf{88.71} & \cellcolor{blue!18}\textbf{83.29} & \multicolumn{1}{c|}{\cellcolor{blue!18}\textbf{95.21}} & \cellcolor{blue!18}\textbf{67.99} & \cellcolor{blue!18}\textbf{57.30} & \multicolumn{1}{c|}{\cellcolor{blue!18}\textbf{82.23}} & \cellcolor{blue!18}\textbf{82.40} & \cellcolor{blue!18}62.81 & \multicolumn{1}{c|}{\cellcolor{blue!18}\textbf{79.67}} & \cellcolor{blue!18}\textbf{77.73} \\ 
\midrule
\multicolumn{1}{c|}{} & \multicolumn{1}{l|}{DeepAligned (AAAI 2021)} & 74.07 & 64.63 & \multicolumn{1}{c|}{88.97} & 49.08 & 37.62 & \multicolumn{1}{c|}{70.50} & 54.50 & 37.96 & \multicolumn{1}{c|}{50.86} & 58.69 \\
\multicolumn{1}{c|}{} & \multicolumn{1}{l|}{MTP-CLNN (ACL 2022)} & 83.26 & 76.20 & \multicolumn{1}{c|}{93.17} & 65.06 & 52.91 & \multicolumn{1}{c|}{80.04} & 74.70 & 54.80 & \multicolumn{1}{c|}{73.35} & 72.61 \\
\multicolumn{1}{c|}{} & \multicolumn{1}{l|}{GCD (CVPR 2022)} & 82.31 & 75.45 & \multicolumn{1}{c|}{92.94} & 69.64 & 58.30 & \multicolumn{1}{c|}{82.17} & 81.60 & 65.90 & \multicolumn{1}{c|}{78.76} & 76.34 \\
\multicolumn{1}{c|}{} & \multicolumn{1}{l|}{ProbNID (ACL 2023)} & 71.56 & 63.25 & \multicolumn{1}{c|}{89.21} & 55.75 & 44.25 & \multicolumn{1}{c|}{74.37} & 54.10 & 38.10 & \multicolumn{1}{c|}{53.70} & 60.48 \\
\multicolumn{1}{c|}{} & \multicolumn{1}{l|}{USNID (TKDE 2023)} & 83.12 & 77.95 & \multicolumn{1}{c|}{94.17} & 65.85 & 56.53 & \multicolumn{1}{c|}{81.94} & 75.76 & 65.45 & \multicolumn{1}{c|}{74.91} & 75.08 \\
\multicolumn{1}{c|}{} & \multicolumn{1}{l|}{SimGCD (ICCV 2023)} & 84.44 & 77.53 & \multicolumn{1}{c|}{93.44} & 69.55 & 57.86 & \multicolumn{1}{c|}{81.71} & 79.80 & 65.19 & \multicolumn{1}{c|}{79.09} & 76.51 \\
\multicolumn{1}{c|}{} & \multicolumn{1}{l|}{CsePL (EMNLP 2023)} & 86.16 & 79.65 & \multicolumn{1}{c|}{94.07} & 71.06 & 60.36 & \multicolumn{1}{c|}{83.22} & 79.47 & 64.92 & \multicolumn{1}{c|}{74.88} & 77.09 \\
\multicolumn{1}{c|}{} & \multicolumn{1}{l|}{ALUP (NAACL 2024)} & 88.40 & 82.44 & \multicolumn{1}{c|}{94.84} & 74.61 & 62.64 & \multicolumn{1}{c|}{84.06} & 82.20 & 64.54 & \multicolumn{1}{c|}{76.58} & 78.92 \\
\multicolumn{1}{c|}{} & \multicolumn{1}{l|}{Loop (ACL 2024)} & 86.58 & 80.67 & \multicolumn{1}{c|}{94.38} & 71.40 & 60.95 & \multicolumn{1}{c|}{83.37} & 82.20 & 66.29 & \multicolumn{1}{c|}{79.10} & 78.33 \\
\multicolumn{1}{c|}{\multirow{-10}{*}{25\%}} & \multicolumn{1}{l|}{\cellcolor{blue!18}\textbf{\MethodName (Ours)}} & \cellcolor{blue!18}\textbf{91.51} & \cellcolor{blue!18}\textbf{87.07} & \multicolumn{1}{c|}{\cellcolor{blue!18}\textbf{96.27}} & \cellcolor{blue!18}\textbf{76.98} & \cellcolor{blue!18}\textbf{66.00} & \multicolumn{1}{c|}{\cellcolor{blue!18}\textbf{85.62}} & \cellcolor{blue!18}\textbf{84.10} & \cellcolor{blue!18}\textbf{71.01} & \multicolumn{1}{c|}{\cellcolor{blue!18}\textbf{80.90}} & \cellcolor{blue!18}\textbf{82.16} \\ 

\midrule

\multicolumn{1}{c|}{} & \multicolumn{1}{l|}{DeepAligned (AAAI 2021)} & 80.70 & 72.56 & \multicolumn{1}{c|}{91.59} & 59.38 & 47.95 & \multicolumn{1}{c|}{76.67} & 74.52 & 57.62 & \multicolumn{1}{c|}{68.28} & 69.92 \\
\multicolumn{1}{c|}{} & \multicolumn{1}{l|}{MTP-CLNN (ACL 2022)} & 86.18 & 80.17 & \multicolumn{1}{c|}{94.30} & 70.97 & 60.17 & \multicolumn{1}{c|}{83.42} & 80.36 & 62.24 & \multicolumn{1}{c|}{76.66} & 77.16 \\
\multicolumn{1}{c|}{} & \multicolumn{1}{l|}{GCD (CVPR 2022)} & 86.53 & 81.06 & \multicolumn{1}{c|}{94.60} & 74.42 & 63.83 & \multicolumn{1}{c|}{84.84} & 85.60 & 72.20 & \multicolumn{1}{c|}{80.12} & 80.36 \\
\multicolumn{1}{c|}{} & \multicolumn{1}{l|}{ProbNID (ACL 2023)} & 82.62 & 75.27 & \multicolumn{1}{c|}{92.72} & 63.02 & 50.42 & \multicolumn{1}{c|}{77.95} & 73.20 & 62.46 & \multicolumn{1}{c|}{74.54} & 72.47 \\
\multicolumn{1}{c|}{} & \multicolumn{1}{l|}{USNID (TKDE 2023)} & 87.22 & 82.87 & \multicolumn{1}{c|}{95.45} & 73.27 & 63.77 & \multicolumn{1}{c|}{85.05} & 82.06 & 71.63 & \multicolumn{1}{c|}{78.77} & 80.01 \\
\multicolumn{1}{c|}{} & \multicolumn{1}{l|}{SimGCD (ICCV 2023)} & 87.24 & 81.65 & \multicolumn{1}{c|}{94.83} & 74.42 & 64.17 & \multicolumn{1}{c|}{85.08} & 82.00 & 70.67 & \multicolumn{1}{c|}{80.44} & 80.06 \\
\multicolumn{1}{c|}{} & \multicolumn{1}{l|}{CsePL (EMNLP 2023)} & 88.66 & 83.14 & \multicolumn{1}{c|}{95.09} & 76.94 & 66.66 & \multicolumn{1}{c|}{85.65} & 85.68 & 71.99 & \multicolumn{1}{c|}{80.28} & 81.57 \\
\multicolumn{1}{c|}{} & \multicolumn{1}{l|}{ALUP (NAACL 2024)} & 90.53 & 84.84 & \multicolumn{1}{c|}{95.97} & 79.45 & 68.78 & \multicolumn{1}{c|}{86.79} & 86.70 & 73.85 & \multicolumn{1}{c|}{81.45} & 83.15 \\
\multicolumn{1}{c|}{} & \multicolumn{1}{l|}{Loop (ACL 2024)} & 90.98 & 85.15 & \multicolumn{1}{c|}{95.59} & 75.06 & 65.70 & \multicolumn{1}{c|}{85.43} & 85.90 & 72.45 & \multicolumn{1}{c|}{80.56} & 81.87 \\
\multicolumn{1}{c|}{\multirow{-10}{*}{50\%}} & \multicolumn{1}{l|}{\cellcolor{blue!18}\textbf{\MethodName (Ours)}} & \cellcolor{blue!18}\textbf{94.53} & \cellcolor{blue!18}\textbf{90.79} & \multicolumn{1}{c|}{\cellcolor{blue!18}\textbf{97.12}} & \cellcolor{blue!18}\textbf{80.26} & \cellcolor{blue!18}\textbf{70.40} & \multicolumn{1}{c|}{\cellcolor{blue!18}\textbf{87.65}} & \cellcolor{blue!18}\textbf{89.40} & \cellcolor{blue!18}\textbf{78.92} & \multicolumn{1}{c|}{\cellcolor{blue!18}\textbf{85.04}} & \cellcolor{blue!18}\textbf{86.01} \\ 

\bottomrule
\end{tabular}%
}
\caption{Main results of \MethodName compared to baseline methods across different datasets and known category ratios (KCR). \MethodName outperforms both standard GCD approaches and the latest LLM-based work Loop \cite{an-etal-2024-generalized}, showing significant improvements especially on the challenging BANKING dataset and with limited known categories. Performance gains are observed across most KCRs, metrics, and datasets.}
\label{tab:main_result}
\end{table*}

Specifically, for each selected ambiguous instance, we query the LLM to identify the most similar category name \& description from those generated in the previous category characterization step. This selected category becomes the positive example, while all other unselected categories serve as negative examples. The corresponding prompt and example are provided in 
Figure \ref{fig:feedback3_prompt_example}.
Then we use a contrastive loss to align the embedding of queried ambiguous data with the embedding of the selected positive category name \& description:
\begin{equation}
\mathcal{L}^{align} = -\frac{1}{N} \sum_{i=1}^N \log \frac{e^{\left(z_i^T d_p / \tau \right)}}{\sum_{j \neq i} e^{{\left(z_i, d_j/ \tau \right)}}}
\end{equation}
where $d_p$ is the embedding of the selected positive category name \& description, $d_j$ is the embedding of other unselected negative category names \& descriptions, and $sim(\cdot, \cdot)$ represents the cosine similarity function.

As a result, the overall loss that leverages labeled data and LLM feedback is formulated as:
\begin{equation}
\mathcal{L} = \mathcal{L}^{ce} + \mathcal{L}^{ncl} + \lambda \mathcal{L}^{align}
\end{equation}
where $\lambda$ is the weight of the alignment loss. 
Note that during model training, we incorporate the supervised cross-entropy loss $\mathcal{L}^{ce}$ on labeled data to enhance model learning from known categories.

\section{Experiments}

\subsection{Experimental Setup}

\textbf{Dataset and Metrics.} We evaluate \MethodName on three standard generalized category discovery benchmarks: BANKING \cite{casanueva-etal-2020-efficient}, CLINC \cite{larson-etal-2019-evaluation} and StackOverflow \cite{xu-etal-2015-short}. We use the same training, validation, and testing splits as previous work \cite{liang-etal-2024-actively, an-etal-2024-generalized}. Descriptions, statistics and setup of all used datasets are provided in Appendix \ref{appendix:dataset}. 
Following \cite{Lin2019DiscoveringNI, zhang-etal-2022-new, liang-etal-2024-actively}, we adopt the following three metrics for evaluation: Clustering Accuracy (ACC), Normalized Mutual Information (NMI) and Adjusted Rand Index (ARI). The specific definitions are provided in Appendix \ref{appendix:metrics}. \\

\noindent \textbf{Baselines.} We compare our model with two types of baselines: (1) Recent SOTA GCD method with LLM feedback: Loop \cite{an-etal-2024-generalized}, ALUP \cite{liang-etal-2024-actively}. (2) Various standard GCD approaches without leveraging LLM feedback: GCD \cite{vaze2022generalized}, SimGCD \cite{wen2023simgcd}, DeepAligned \cite{zhang2021discovering}, MTP-CLNN \cite{zhang-etal-2022-new}, ProbNID \cite{zhou-etal-2023-probabilistic}, USNID \cite{zhang2023clustering}, CsePL \cite{liang-liao-2023-clusterprompt}. \textit{gpt-4o-mini} is used as the default LLM to acquire LLM feedback.\\


\noindent \textbf{Implementation Details.} Following \cite{an-etal-2024-generalized}, we take the BERT-based-uncased model \cite{wolf-etal-2020-transformers} as our base model and pre-train it first on both labeled and unlabeled data from the current dataset using cross-entropy loss and masked language modeling loss \cite{Devlin2019BERTPO}. 
We use the \textit{[CLS]} token as initial text features for clustering. For contrastive learning, we employ a two-layer MLP to project the 768-d initial features into a 128-d space. A complete list of default hyper-parameters is provided in Appendix \ref{appendix:hyperparameters}. To reduce the computing and query cost, we follow the practice of \cite{an-etal-2024-generalized}: mine ambiguous data and update the query set every 5 epochs, and repeat the described approach 5 times.

\vspace{0.5cm}
\subsection{Main Results}
We summarize our main results under different known category ratios in Table \ref{tab:main_result} and describe our key findings below:
\vspace{0.5cm}

\noindent \textbf{Compare with standard GCD approaches.} It can be observed that our \MethodName consistently achieves stronger performance than standard GCD approaches. For example, on the most challenging BANKING dataset with 25\% known category ratio, \MethodName significantly outperforms CsePL \cite{liang-liao-2023-clusterprompt} by 5.35\%/7.42\%, and SimGCD \cite{wen2023simgcd} by 7.07\%/9.54\% in terms of ACC/ARI. These substantial improvements validate the effectiveness of our overall pipeline of active generalized category discovery from diverse LLM feedback.
\vspace{0.2cm}

\noindent \textbf{Compare with state-of-the-art LLM-based GCD.} \MethodName outperforms the recent state-of-the-art GCD method with LLM feedback, Loop \cite{an-etal-2024-generalized} and ALUP \cite{liang-etal-2024-actively}, across most settings and datasets. This verifies our motivation that multiple diverse feedback can be leveraged to effectively boost GCD model performance. Noteworthy is that, \MethodName can offer remarkable improvements over Loop even with extremely limited known categories: +9.38\% ACC on BANKING and +3.5\% ACC on StackOverflow with 5\% known category ratio. Note that we re-ran the released code for GCD, SimGCD, and Loop to obtain their results, while the results for other baselines were retrieved from \cite{liang-etal-2024-actively}.
\vspace{0.2cm}

\noindent \textbf{Superior Overall Performance.} The performance gain of \MethodName is generally maintained across different known category ratios, different metrics and datasets, showing the robustness of our approach. Furthermore, on the CLINC dataset, \MethodName achieves over 90\% ACC with 25\% KCR, demonstrating its potential for high-accuracy category discovery in these domains.

\begin{table}[!t]
\centering
\resizebox{\columnwidth}{!}{%
\begin{tabular}{llccccc}
\toprule
\textbf{Datasets} & \textbf{BANK} & \textbf{CLINC} & \textbf{STACK} & \textbf{Average} \\ \midrule
\multicolumn{5}{l}{\textit{Instance-Instance LLM Feedback: Similar Instance Selection}} \\
Random Selection & 0.037 & 0.070 & 0.060 & 0.056 \\
Semantic Nearest & 0.273 & 0.377 & 0.243 & 0.298 \\
Naive LLM Selection & 0.337 & 0.510 & 0.423 & 0.423 \\
{\cellcolor{blue!18}w. In-Context Demon} & {\cellcolor{blue!18}0.433} & {\cellcolor{blue!18}0.543} & {\cellcolor{blue!18}0.483} & {\cellcolor{blue!18}0.487} \\
{\cellcolor{blue!18}w. Filtering} & {\cellcolor{blue!18}0.466} & {\cellcolor{blue!18}0.562} & {\cellcolor{blue!18}0.563} & {\cellcolor{blue!18}0.530} \\ \midrule \midrule
\textbf{Sample Types} & \multicolumn{2}{c}{\textbf{Hardest Samples}} & \multicolumn{2}{c}{\textbf{Random Samples}} \\ \midrule
\textbf{Datasets} & \textbf{BANK} & \textbf{CLINC} & \textbf{BANK} & \textbf{CLINC} \\ \midrule
\multicolumn{5}{l}{\textit{Cluster-Instance LLM Feedback: Category Selection}} \\
Naive LLM Selection & 0.517 & 0.757 & 0.707 & 0.807 \\
{\cellcolor{blue!18}w. In-Context Demon} & {\cellcolor{blue!18}0.530} & {\cellcolor{blue!18}0.797} & {\cellcolor{blue!18}0.717} & {\cellcolor{blue!18}0.860} \\
{\cellcolor{blue!18}w. Filtering} & {\cellcolor{blue!18}0.557} & {\cellcolor{blue!18}0.859} & {\cellcolor{blue!18}0.847} & {\cellcolor{blue!18}0.943} \\ \bottomrule
\end{tabular}%
}
\caption{LLM feedback quality investigation (accuracy). Naively prompting LLM, as done in previous work, yields unsatisfactory results, though still much higher than random and semantic nearest selection. LLM feedback quality can be greatly enhanced with in-context demonstrations and filtering. More details in Section \ref{sec:llm_feedback_quality_investigation}.}
\label{tab:feedback_quality_investigation}
\end{table}

\begin{table*}[tbh]
\centering
\resizebox{0.95\textwidth}{!}{%
\begin{tabular}{llccccccccccc}
\toprule
 & \multicolumn{3}{c}{\textbf{CLINC}} & \multicolumn{3}{c}{\textbf{BANKING}} & \multicolumn{3}{c}{\textbf{Stackoverflow}} & \multicolumn{1}{l}{} \\ \midrule
\multicolumn{1}{l|}{\textbf{Method}} & \textbf{ACC} & \textbf{ARI} & \multicolumn{1}{c|}{\textbf{NMI}} & \textbf{ACC} & \textbf{ARI} & \multicolumn{1}{c|}{\textbf{NMI}} & \textbf{ACC} & \textbf{ARI} & \multicolumn{1}{c|}{\textbf{NMI}} & \textbf{Average} \\ \midrule

\multicolumn{1}{l|}{\cellcolor{blue!18}\MethodName} & \cellcolor{blue!18}91.51 & \cellcolor{blue!18}87.07 & \multicolumn{1}{c|}{\cellcolor{blue!18}96.27} & \cellcolor{blue!18}76.98 & \cellcolor{blue!18}66.00 & \multicolumn{1}{c|}{\cellcolor{blue!18}85.62} & \cellcolor{blue!18}84.10 &\cellcolor{blue!18}71.01 & \multicolumn{1}{c|}{\cellcolor{blue!18}80.90} & \cellcolor{blue!18}82.16 \\
\multicolumn{1}{l|}{\cellcolor{blue!18}\MethodName w/o Filtering \& Demo} & \cellcolor{blue!18}91.02 & \cellcolor{blue!18}86.32 & \multicolumn{1}{c|}{\cellcolor{blue!18}96.06} & \cellcolor{blue!18}76.14 & \cellcolor{blue!18}65.60 & \multicolumn{1}{c|}{\cellcolor{blue!18}85.45} & \cellcolor{blue!18}83.40 & \cellcolor{blue!18}68.93 & \multicolumn{1}{c|}{\cellcolor{blue!18}80.10} & \cellcolor{blue!18}81.45 \\ \midrule
\multicolumn{1}{l|}{- Cross-Entropy Loss} & 89.73 & 84.72 & \multicolumn{1}{c|}{95.57} & 72.21 & 62.22 & \multicolumn{1}{c|}{84.12} & 82.30 & 67.25 & \multicolumn{1}{c|}{79.13} & 79.69 \\
\multicolumn{1}{l|}{- Neighborhood Contrastive Learning} & 79.20 & 71.90 & \multicolumn{1}{c|}{92.32} & 67.40 & 56.13 & \multicolumn{1}{c|}{81.77} & 72.40 & 56.81 & \multicolumn{1}{c|}{71.53} & 72.16 \\
\multicolumn{1}{l|}{- Instance-Instance LLM Feedback} & 88.04 & 83.90 & \multicolumn{1}{c|}{95.54} & 71.95 & 63.44 & \multicolumn{1}{c|}{84.78} & 80.80 & 65.34 & \multicolumn{1}{c|}{78.31} & 79.12 \\
\multicolumn{1}{l|}{- Cluster-Instance LLM Feedback} & 88.18 & 82.98 & \multicolumn{1}{c|}{95.00} & 72.56 & 62.23 & \multicolumn{1}{c|}{83.86} & 79.60 & 62.71 & \multicolumn{1}{c|}{76.90} & 78.22 \\ \bottomrule
\end{tabular}%
}
\caption{Ablation study of the different components of \MethodName. Each component contributes to the final performance, with Cluster-Instance LLM Feedback and Neighborhood Contrastive Learning being particularly impactful. Instance-Level LLM Feedback refers to Similar Instance Selection. Cluster-Instance LLM Feedback refers to both Category Characterization and Pseudo Category Selection and Alignment.}
\label{tab:ablation_component}
\end{table*}

\begin{table*}[tbh]
\centering
\resizebox{0.95\textwidth}{!}{%
\begin{tabular}{@{}cccccccccccc@{}}
\toprule
\multicolumn{1}{l}{} & \multicolumn{1}{l}{} & \multicolumn{3}{c}{\textbf{CLINC}} & \multicolumn{3}{c}{\textbf{BANKING}} & \multicolumn{3}{c}{\textbf{Stackoverflow}} & \multicolumn{1}{l}{} \\ \midrule
\textbf{LLMs} & \multicolumn{1}{c|}{\textbf{Cost}} & \textbf{ACC} & \textbf{ARI} & \multicolumn{1}{c|}{\textbf{NMI}} & \textbf{ACC} & \textbf{ARI} & \multicolumn{1}{c|}{\textbf{NMI}} & \textbf{ACC} & \textbf{ARI} & \multicolumn{1}{c|}{\textbf{NMI}} & \textbf{Average} \\ \midrule
gpt-3.5-turbo & \multicolumn{1}{c|}{\$0.5/ \$1.5} & 89.29 & 84.47 & \multicolumn{1}{c|}{95.54} & 73.38 & 64.01 & \multicolumn{1}{c|}{85.20} & 81.40 & 68.91 & \multicolumn{1}{c|}{80.24} & 80.27 \\
gpt-4o-mini & \multicolumn{1}{c|}{\$0.15 / \$0.6} & 91.02 & 86.32 & \multicolumn{1}{c|}{96.06} & 76.14 & 65.60 & \multicolumn{1}{c|}{85.45} & 83.40 & 68.93 & \multicolumn{1}{c|}{80.10} & 81.45 \\
gpt-4o & \multicolumn{1}{c|}{\$2.5 / \$10} & 90.89 & 86.36 & \multicolumn{1}{c|}{96.12} & 76.27 & 65.36 & \multicolumn{1}{c|}{85.44} & 82.40 & 67.20 & \multicolumn{1}{c|}{79.45} & 81.05 \\
DeepSeek-V3 & \multicolumn{1}{c|}{\$1.25 / \$1.25} & 91.91 & 87.28 & \multicolumn{1}{c|}{96.28} & 76.07 & 65.93 & \multicolumn{1}{c|}{85.65} & 83.40 & 69.20 & \multicolumn{1}{c|}{80.02} & 81.75 \\
Qwen-2.5-72B & \multicolumn{1}{c|}{\$1.2 / \$1.2} & 91.16 & 86.90 & \multicolumn{1}{c|}{96.15} & 74.90 & 64.61 & \multicolumn{1}{c|}{85.27} & 82.30 & 67.42 & \multicolumn{1}{c|}{79.39} & 80.90 \\ 
Llama-3.3-70B & \multicolumn{1}{c|}{\$0.88 / \$0.88} & 90.84 & 86.19 & \multicolumn{1}{c|}{96.09} & 74.38 & 64.43 & \multicolumn{1}{c|}{85.17} & 80.70 & 66.99 & \multicolumn{1}{c|}{79.17} & 80.44 \\
Qwen-2.5-7B & \multicolumn{1}{c|}{\$0.3 / \$0.3} & 90.13 & 85.82 & \multicolumn{1}{c|}{95.89} & 73.21 & 62.89 & \multicolumn{1}{c|}{84.22} & 80.60 & 66.62 & \multicolumn{1}{c|}{79.31} & 79.85 \\ 
Llama-3.2-3B & \multicolumn{1}{c|}{\$0.06 / \$0.06} & 88.80 & 83.07 & \multicolumn{1}{c|}{94.89} & 72.60 & 62.00 & \multicolumn{1}{c|}{84.27} & 80.10 & 66.50 & \multicolumn{1}{c|}{78.90} & 79.01 \\

\bottomrule
\end{tabular}%
}
\caption{Different variants of LLMs. Open-source models, particularly \textit{DeepSeek-V3}, can compete with closed-source alternatives. \textit{gpt-4o-mini} offers a good balance between cost and performance, achieving the second-highest scores at a relatively low cost. Cost: pricing per 1M input tokens and output tokens.}
\label{tab:variants_llm}
\end{table*}



\section{Ablation Study and Analysis}

\label{sec:llm_feedback_quality_investigation}
\subsection{LLM Feedback Quality Investigation}

In this section, we present our pilot investigation into the quality of diverse LLM feedback and introduce simple strategies to enhance its quality. We conduct experiments on three standard GCD datasets: BANKING \cite{casanueva-etal-2020-efficient}, CLINC \cite{larson-etal-2019-evaluation}, and StackOverflow \cite{xu-etal-2015-short}. Accuracy is used as the evaluation metric, calculated by comparing ground-truth answers with predictions. To better reflect real-world use cases, we evaluate performance on the 300 most challenging or ambiguous samples from each dataset. (Details on ambiguous data mining are provided in Section \ref{sec:ambiguous_data_mining}).

Table \ref{tab:feedback_quality_investigation} summarizes the results of our investigation into two types of LLM feedback quality:
(1) Instance-Instance LLM Feedback: Similar Instance Selection \cite{an-etal-2024-generalized}. For each ambiguous instance, we randomly sample $M$ instances from each of its closest $M$ clusters as candidates and then query the LLM to select the instance most similar to the ambiguous data as the positive instance.
(2) Cluster-Instance LLM Feedback: Category Selection. For each ambiguous instance, we query the LLM to identify the most similar category from a given list of candidate categories. Our findings show that naively prompting the LLM, as done in previous work \cite{liang-etal-2024-actively, an-etal-2024-generalized}, leads to unsatisfactory results, achieving an average accuracy of only 0.487 for similar instance selection. While this is still much higher than random selection or semantic nearest-neighbor selection and can aid model learning, improving feedback quality is more beneficial for model training. To address this, we adopt two simple strategies to enhance LLM feedback quality and mitigate noise. When querying the LLM, we incorporate in-context demonstrations from known categories and ask the LLM not only to provide an answer but also to output its confidence in the response, and we then filter out low-confidence answers. As shown in Table \ref{tab:feedback_quality_investigation}, these strategies substantially improve LLM feedback quality for both types of feedback.



\subsection{Effectiveness of Each Component}

To show the effectiveness of each component in \MethodName, we measure the performance after removing different parts across the three benchmarking datasets with 25\% known category ratio in Table \ref{tab:ablation_component}. We observe that the performance decreases after stripping each component, suggesting that all components in \MethodName contribute to the final performance. The performance of \MethodName drops most significantly after removing Neighborhood Contrastive Learning across all three datasets. This justifies the importance of leveraging LLMs to identify similar instances among ambiguous data points and refining embeddings through neighborhood contrastive learning. Besides, it can be seen that in-context demonstration and filtering low-confidence LLM feedback indeed help boost model performance, which aligns with our investigation and findings in Section \ref{sec:llm_feedback_quality_investigation}.

Interestingly, compared to the Instance-Instance LLM Feedback - Similar Instance Selection, model performance drops more after removing Cluster-Instance LLM Feedback on most datasets and metrics. Specifically, removing Cluster-Instance LLM Feedback results in a 3.23\% decrease in ACC, while removing Instance-Instance LLM Feedback leads to a 2.33\% decrease. This finding underscores the benefits of aligning instance embeddings with corresponding LLM-selected category descriptions, which fosters improved representation learning by considering category-instance relationships. Not surprisingly, the removal of Cross-Entropy Loss also leads to a noticeable performance drop, particularly on the BANKING and StackOverflow datasets. This validates the importance of supervised learning on labeled data to enhance model performance on known categories and establish a strong foundation for subsequent LLM-enhanced learning processes. These results demonstrate that each component of \MethodName plays a crucial role in its overall performance.

\subsection{Different Variants of LLMs}
We investigate the performance impact of using different variants of LLMs in Table \ref{tab:variants_llm}. Note that the pricing for GPT-series models is sourced from OpenAI\footnote{https://platform.openai.com/docs/pricing}, while the pricing for all open-source models is sourced from Together.AI\footnote{https://www.together.ai/pricing}. It can be observed that DeepSeek-V3 achieves the highest overall average performance (81.75\%), closely followed by the closed-source gpt-4o-mini (81.45\%) and gpt-4o (81.05\%), demonstrating that open-source models can compete with proprietary alternatives. 
Notably, gpt-4o-mini offers the best balance between cost and effectiveness, delivering competitive performance at a significantly lower cost (\$0.15 / \$0.6). In dataset-specific trends, DeepSeek-V3 excels in CLINC with the highest accuracy (91.91\%), while gpt-4o-mini leads in BANKING (76.14\% ACC), and both DeepSeek-V3 and gpt-4o-mini perform best on Stackoverflow (83.40\% ACC). These results suggest that while OpenAI's closed-source models remain strong, open-source alternatives like DeepSeek-V3 are closing the gap and providing viable, high-performing options.


\begin{figure}[!tb]
    \centering
    \includegraphics[width=\columnwidth]{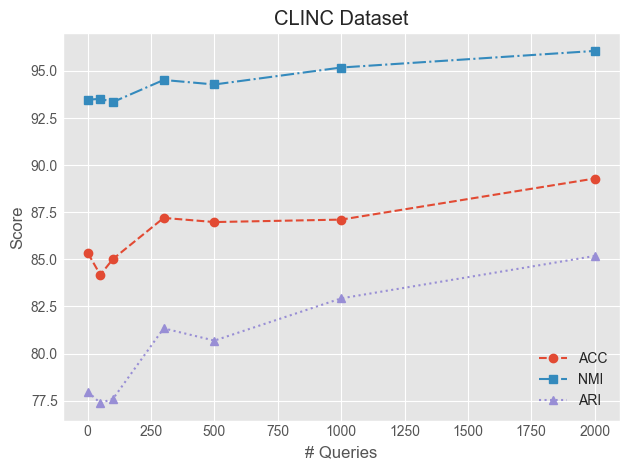} 
    \caption{Influence of the number of query samples. Increasing the number of query samples generally leads to better performance.}
    \label{fig:num_query_samples}
\end{figure}

\subsection{Influence of Query Sample Number}
We study the influence of the number of query samples in Figure \ref{fig:num_query_samples}. We can observe that increasing the number of query samples generally leads to better performance, as more LLM feedback signals are available. For instance, on the  CLINC dataset, the ARI increases from 77.5\% to 85.0\% as the number of query samples increases from 0 to 2000, and the trend shows that more performance gain can be obtained with more query samples. Yet, the performance gain starts to saturate on the BANKING dataset as the number of query samples reaches 500, we hypothesize this is because the BANKING dataset is more challenging and distinguishing ambiguous samples and categories becomes increasingly difficult as the number of samples increases.

\begin{table}[!t]
\resizebox{\columnwidth}{!}{%
\begin{tabular}{@{}ccccccc@{}}
\toprule
\textbf{Dataset} & \multicolumn{1}{l}{} & \textbf{CLINC} & \multicolumn{1}{l}{} & \multicolumn{1}{l}{} & \textbf{BANKING} & \multicolumn{1}{l}{} \\ \midrule
\textbf{Clustering Algorithm} & \textbf{ACC} & \textbf{NMI} & \textbf{ARI} & \textbf{ACC} & \textbf{NMI} & \textbf{ARI} \\
\midrule
K-Means++ & 91.16 & 95.85 & 85.99 & 75.88 & 85.62 & 65.51 \\
HDBSCAN & 90.53 & 95.79 & 85.51 & 75.36 & 85.17 & 64.36 \\
GMM & 89.82 & 95.67 & 84.92 & 75.03 & 85.49 & 64.50 \\ \bottomrule
\end{tabular}%
}
\caption{Results with other clustering algorithms.}
\label{tab:clustering_algorithms}
\end{table}

\begin{table}[!t]
\resizebox{\columnwidth}{!}{%
\begin{tabular}{@{}ccccccc@{}}
\toprule
\textbf{Epoch} & \textbf{All} & \textbf{Head} & \textbf{Middle} & \textbf{Tail} & \textbf{Known} & \textbf{Novel} \\ \midrule
\# Instances Range & 168$\sim$32 & 168$\sim$134 & 133$\sim$106 & 104$\sim$32 & - & - \\ \midrule
0 & 52.44 & 56.20 & 45.3 & 55.56 & 67.5 & 47.5 \\
5 & 65.13 & 71.5 & 60 & 63.98 & 78.16 & 60.86 \\
10 & 71.98 & 72.6 & 75.6 & 68.06 & 82.37 & 68.58 \\
15 & 74.87 & 75.8 & 76.5 & 72.5 & 83.68 & 71.98 \\
20 & 75.06 & 71.6 & 75.4 & 77.96 & 82.63 & 72.59 \\
25 & 76.14 & 79.1 & 76.2 & 73.33 & 80.92 & 74.57 \\ \bottomrule
\end{tabular}%
}
\caption{Detailed breakdown of accuracy results on different types of classes on imbalanced dataset.}
\label{tab:impact_of_imbalance}
\end{table}

\subsection{Results with Other Clustering Algorithms}

Our method is also compatible with other common clustering algorithms such as GMM (Gaussian Mixuture Models) or clustering algorithms with automatic K (category number) estimation (e.g., DBSCAN, HDBSCAN). We have added experiments using HDBSCAN and GMM in our \MethodName method. The results in Table \ref{tab:clustering_algorithms} show that all three clustering algorithms achieve similar performance, and our method with K-Means++ achieves competitive performance, likely due to the refined representation space learned via LLM feedback and contrastive objectives.

\subsection{Impact of class imbalance}

This section adds a detailed breakdown of accuracy results on different kinds of classes on the imbalanced dataset BANKING (Known Category Ratio = 25\%) across different training epochs. Head/Tail denotes the head/tail one-third of classes with the most/least training instances, ranging from 168\texttt{\string~}134/104\texttt{\string~}32. We can see that LLM feedback can effectively improve the performance of both Head and Tail classes throughout the training process. More interestingly, we observe that the performance improvement on novel classes (+26.07\%) is more obvious than the performance improvement on known classes (+14.42\%), especially in the later training phase, as more ambiguous data can be mined from novel classes and can receive LLM feedback for refinement.

\subsection{Computational Cost}

In response, we have added a new table summarizing the actual token usage, total \$ cost, and wall-clock LLM query time for \MethodName across all datasets. These measurements were collected using gpt-4o-mini under the default query budget of 500 ambiguous samples. As shown below, a full run requires 1.55M–5.88M tokens depending on dataset, which corresponds to \$0.25-\$0.91 per training run, and all LLM querying completes in under 15 minutes. These results demonstrate that \MethodName is extremely cost-efficient and practical for real-world deployment, with total cost less than \$1 per run when using gpt-4o-mini.

\begin{table}[!t]
\resizebox{\columnwidth}{!}{%
\begin{tabular}{@{}lccccc@{}}
\toprule
Dataset & \textbf{\begin{tabular}[c]{@{}c@{}}Input \\ Tokens (M)\end{tabular}} & \textbf{\begin{tabular}[c]{@{}c@{}}Output \\ Tokens (M)\end{tabular}} & \textbf{\begin{tabular}[c]{@{}c@{}}Total \\ Tokens (M)\end{tabular}} & \textbf{\begin{tabular}[c]{@{}c@{}}Total \\ Cost (\$)\end{tabular}} & \textbf{\begin{tabular}[c]{@{}c@{}}Query \\ Time (min)\end{tabular}} \\ \midrule
BANKING & 3.63178 & 0.05924 & 3.69102 & 0.58031 & 12.32 \\
CLINC & 5.81621 & 0.06585 & 0.58821 & 0.91194 & 14.80 \\
StackOverflow & 1.50342 & 0.05270 & 1.55612 & 0.25713 & 11.00 \\ \bottomrule
\end{tabular}%
}
\caption{Computational cost on three evaluated datasets.}
\label{tab:cost}
\end{table}

\subsection{Robustness Results}
To demonstrate the robustness of our methods, we ran our method three times on both the BANKING and CLINC benchmarks and report the mean and standard deviation below. The low standard deviations (e.g., ACC standard deviation less than 0.64 on banking dataset) demonstrate that our approach yields robust performance across runs and still consistently achieves SOTA performance despite the inherent noise and stochasticity of LLM feedback.

\begin{table}[!t]
\resizebox{\columnwidth}{!}{%
\begin{tabular}{@{}ccccccc@{}}
\toprule
\multicolumn{1}{l}{} & \multicolumn{1}{l}{} & \textbf{BANKING} & \multicolumn{1}{l}{} & \multicolumn{1}{l}{} & \textbf{CLINC} & \multicolumn{1}{l}{} \\ \midrule
Run & ACC & NMI & ARI & ACC & NMI & ARI \\ \midrule
1 & 76.98 & 85.62 & 66.00 & 91.51 & 96.27 & 87.07 \\
2 & 75.88 & 85.62 & 65.51 & 91.16 & 95.85 & 85.99 \\
3 & 75.88 & 84.99 & 64.59 & 90.44 & 95.85 & 85.64 \\
Avg & 76.25 & 85.41 & 65.37 & 91.04 & 95.99 & 86.23 \\
\textbf{Std} & \textbf{0.635} & \textbf{0.364} & \textbf{0.716} & \textbf{0.546} & \textbf{0.242} & \textbf{0.745} \\ \bottomrule
\end{tabular}%
}
\caption{Mean performance and standard deviation.}
\label{tab:robustness_results}
\end{table}

\subsection{Comparison with LLM-native Baselines}

In this section, we have added GPT-4o and GPT-4o with 10 in-context demonstrations as two additional LLM-native baselines, and the accuracy results are shown in the table below. Even when we provide all category names and corresponding in-context examples to GPT-4o (KCR = 100\%), our \MethodName still consistently delivers improvements over these baselines while assuming only KCR = 25\% and 50\%, demonstrating the effectiveness of our method in combining both SLM training and LLM feedback.

\begin{table}[!t]
\centering
\resizebox{\columnwidth}{!}{%
\begin{tabular}{@{}lcc@{}}
\toprule
 & \textbf{BANKING} & \textbf{CLINC} \\ \midrule
GPT4o (KCR=100\%) & 70.67 & 80.67 \\
w. In-Context Demonstration & 71.67 & 86.00 \\
DeLFGCD (KCR=25\%) & 76.98 & 91.51 \\
DeLFGCD (KCR=50\%) & 80.26 & 94.53 \\ \bottomrule
\end{tabular}%
}
\caption{Comparison with LLM-native baselines.}
\label{tab:llm_native_baseline}
\end{table}

\section{Conclusion}

This paper introduces \MethodName, a holistic framework that leverages diverse and quality-enhanced LLM feedback for generalized category discovery. Our approach addresses key limitations of existing methods, including insufficient supervision, lack of self-correction mechanisms for ambiguous data, and underutilization of semantic meanings in discovered categories. We integrate both instance-level and cluster-level LLM feedback into a contrastive learning framework, while aligning the embeddings of ambiguous instances with LLM-generated and selected category descriptions. Using three real-world datasets, we report new state-of-the-art results with  \MethodName over existing methods across various supervision setups. We provide comprehensive ablation studies and analyses to understand each component in our framework.

\section*{Limitations}
Our current framework is designed for textual data, which limits its applicability to other domains. In the future, we plan to extend it to vision and multimodal domains, exploring learning from multimodal large language models. Additionally, when using external LLMs, data privacy and security remain critical concerns that require ongoing vigilance. Leveraging open-source models such as DeepSeek, LLaMa could help mitigate these risks. 


\newpage
\bibliography{custom}

@inproceedings{chen2025semi,
  title={Semi-supervised node importance estimation with informative distribution modeling for uncertainty regularization},
  author={Chen, Yankai and Wang, Taotao and Fang, Yixiang and Xiao, Yunyu},
  booktitle={Proceedings of the ACM on Web Conference 2025},
  pages={3108--3118},
  year={2025}
}

@inproceedings{vaze2022generalized,
  title={Generalized category discovery},
  author={Vaze, Sagar and Han, Kai and Vedaldi, Andrea and Zisserman, Andrew},
  booktitle={Proceedings of the IEEE/CVF Conference on Computer Vision and Pattern Recognition},
  pages={7492--7501},
  year={2022}
}

@inproceedings{pu2023dynamic,
  title={Dynamic conceptional contrastive learning for generalized category discovery},
  author={Pu, Nan and Zhong, Zhun and Sebe, Nicu},
  booktitle={Proceedings of the IEEE/CVF conference on computer vision and pattern recognition},
  pages={7579--7588},
  year={2023}
}

@inproceedings{liang-etal-2024-actively,
    title = "Actively Learn from {LLM}s with Uncertainty Propagation for Generalized Category Discovery",
    author = "Liang, Jinggui  and
      Liao, Lizi  and
      Fei, Hao  and
      Li, Bobo  and
      Jiang, Jing",
    editor = "Duh, Kevin  and
      Gomez, Helena  and
      Bethard, Steven",
    booktitle = "Proceedings of the 2024 Conference of the North American Chapter of the Association for Computational Linguistics: Human Language Technologies (Volume 1: Long Papers)",
    month = jun,
    year = "2024",
    address = "Mexico City, Mexico",
    publisher = "Association for Computational Linguistics",
    url = "https://aclanthology.org/2024.naacl-long.434",
    doi = "10.18653/v1/2024.naacl-long.434",
    pages = "7845--7858",
}

@inproceedings{an-etal-2024-generalized,
    title = "Generalized Category Discovery with Large Language Models in the Loop",
    author = "An, Wenbin  and
      Shi, Wenkai  and
      Tian, Feng  and
      Lin, Haonan  and
      Wang, QianYing  and
      Wu, Yaqiang  and
      Cai, Mingxiang  and
      Wang, Luyan  and
      Chen, Yan  and
      Zhu, Haiping  and
      Chen, Ping",
    editor = "Ku, Lun-Wei  and
      Martins, Andre  and
      Srikumar, Vivek",
    booktitle = "Findings of the Association for Computational Linguistics ACL 2024",
    month = aug,
    year = "2024",
    address = "Bangkok, Thailand and virtual meeting",
    publisher = "Association for Computational Linguistics",
    url = "https://aclanthology.org/2024.findings-acl.512",
    pages = "8653--8665",
}

@inproceedings{casanueva-etal-2020-efficient,
    title = "Efficient Intent Detection with Dual Sentence Encoders",
    author = "Casanueva, I{\~n}igo  and
      Tem{\v{c}}inas, Tadas  and
      Gerz, Daniela  and
      Henderson, Matthew  and
      Vuli{\'c}, Ivan",
    editor = "Wen, Tsung-Hsien  and
      Celikyilmaz, Asli  and
      Yu, Zhou  and
      Papangelis, Alexandros  and
      Eric, Mihail  and
      Kumar, Anuj  and
      Casanueva, I{\~n}igo  and
      Shah, Rushin",
    booktitle = "Proceedings of the 2nd Workshop on Natural Language Processing for Conversational AI",
    month = jul,
    year = "2020",
    address = "Online",
    publisher = "Association for Computational Linguistics",
    url = "https://aclanthology.org/2020.nlp4convai-1.5",
    doi = "10.18653/v1/2020.nlp4convai-1.5",
    pages = "38--45",
}

@inproceedings{larson-etal-2019-evaluation,
    title = "An Evaluation Dataset for Intent Classification and Out-of-Scope Prediction",
    author = "Larson, Stefan  and
      Mahendran, Anish  and
      Peper, Joseph J.  and
      Clarke, Christopher  and
      Lee, Andrew  and
      Hill, Parker  and
      Kummerfeld, Jonathan K.  and
      Leach, Kevin  and
      Laurenzano, Michael A.  and
      Tang, Lingjia  and
      Mars, Jason",
    editor = "Inui, Kentaro  and
      Jiang, Jing  and
      Ng, Vincent  and
      Wan, Xiaojun",
    booktitle = "Proceedings of the 2019 Conference on Empirical Methods in Natural Language Processing and the 9th International Joint Conference on Natural Language Processing (EMNLP-IJCNLP)",
    month = nov,
    year = "2019",
    address = "Hong Kong, China",
    publisher = "Association for Computational Linguistics",
    url = "https://aclanthology.org/D19-1131",
    doi = "10.18653/v1/D19-1131",
    pages = "1311--1316",
}

@inproceedings{xu-etal-2015-short,
    title = "Short Text Clustering via Convolutional Neural Networks",
    author = "Xu, Jiaming  and
      Wang, Peng  and
      Tian, Guanhua  and
      Xu, Bo  and
      Zhao, Jun  and
      Wang, Fangyuan  and
      Hao, Hongwei",
    editor = "Blunsom, Phil  and
      Cohen, Shay  and
      Dhillon, Paramveer  and
      Liang, Percy",
    booktitle = "Proceedings of the 1st Workshop on Vector Space Modeling for Natural Language Processing",
    month = jun,
    year = "2015",
    address = "Denver, Colorado",
    publisher = "Association for Computational Linguistics",
    url = "https://aclanthology.org/W15-1509",
    doi = "10.3115/v1/W15-1509",
    pages = "62--69",
}

@inproceedings{zhang-etal-2022-new,
    title = "New Intent Discovery with Pre-training and Contrastive Learning",
    author = "Zhang, Yuwei  and
      Zhang, Haode  and
      Zhan, Li-Ming  and
      Wu, Xiao-Ming  and
      Lam, Albert",
    editor = "Muresan, Smaranda  and
      Nakov, Preslav  and
      Villavicencio, Aline",
    booktitle = "Proceedings of the 60th Annual Meeting of the Association for Computational Linguistics (Volume 1: Long Papers)",
    month = may,
    year = "2022",
    address = "Dublin, Ireland",
    publisher = "Association for Computational Linguistics",
    url = "https://aclanthology.org/2022.acl-long.21",
    doi = "10.18653/v1/2022.acl-long.21",
    pages = "256--269",
}

@article{Lin2019DiscoveringNI,
  title={Discovering New Intents via Constrained Deep Adaptive Clustering with Cluster Refinement},
  author={Ting-En Lin and Hua Xu and Hanlei Zhang},
  journal={ArXiv},
  year={2019},
  volume={abs/1911.08891},
  url={https://api.semanticscholar.org/CorpusID:208176145}
}

@inproceedings{wolf-etal-2020-transformers,
    title = "Transformers: State-of-the-Art Natural Language Processing",
    author = "Wolf, Thomas  and
      Debut, Lysandre  and
      Sanh, Victor  and
      Chaumond, Julien  and
      Delangue, Clement  and
      Moi, Anthony  and
      Cistac, Pierric  and
      Rault, Tim  and
      Louf, Remi  and
      Funtowicz, Morgan  and
      Davison, Joe  and
      Shleifer, Sam  and
      von Platen, Patrick  and
      Ma, Clara  and
      Jernite, Yacine  and
      Plu, Julien  and
      Xu, Canwen  and
      Le Scao, Teven  and
      Gugger, Sylvain  and
      Drame, Mariama  and
      Lhoest, Quentin  and
      Rush, Alexander",
    editor = "Liu, Qun  and
      Schlangen, David",
    booktitle = "Proceedings of the 2020 Conference on Empirical Methods in Natural Language Processing: System Demonstrations",
    month = oct,
    year = "2020",
    address = "Online",
    publisher = "Association for Computational Linguistics",
    url = "https://aclanthology.org/2020.emnlp-demos.6",
    doi = "10.18653/v1/2020.emnlp-demos.6",
    pages = "38--45",
}

@inproceedings{zou-caragea-2023-jointmatch,
    title = "{J}oint{M}atch: A Unified Approach for Diverse and Collaborative Pseudo-Labeling to Semi-Supervised Text Classification",
    author = "Zou, Henry  and
      Caragea, Cornelia",
    editor = "Bouamor, Houda  and
      Pino, Juan  and
      Bali, Kalika",
    booktitle = "Proceedings of the 2023 Conference on Empirical Methods in Natural Language Processing",
    month = dec,
    year = "2023",
    address = "Singapore",
    publisher = "Association for Computational Linguistics",
    url = "https://aclanthology.org/2023.emnlp-main.451",
    doi = "10.18653/v1/2023.emnlp-main.451",
    pages = "7290--7301",
}

@inproceedings{an2023generalized,
  title={Generalized category discovery with decoupled prototypical network},
  author={An, Wenbin and Tian, Feng and Zheng, Qinghua and Ding, Wei and Wang, QianYing and Chen, Ping},
  booktitle={Proceedings of the AAAI Conference on Artificial Intelligence},
  volume={37},
  pages={12527--12535},
  year={2023}
}

@inproceedings{tang-etal-2023-rsvp,
    title = "{RSVP}: Customer Intent Detection via Agent Response Contrastive and Generative Pre-Training",
    author = "Tang, Yu-Chien  and
      Wang, Wei-Yao  and
      Yen, An-Zi  and
      Peng, Wen-Chih",
    editor = "Bouamor, Houda  and
      Pino, Juan  and
      Bali, Kalika",
    booktitle = "Findings of the Association for Computational Linguistics: EMNLP 2023",
    month = dec,
    year = "2023",
    address = "Singapore",
    publisher = "Association for Computational Linguistics",
    url = "https://aclanthology.org/2023.findings-emnlp.698",
    doi = "10.18653/v1/2023.findings-emnlp.698",
    pages = "10400--10412",
}

@inproceedings{gong-etal-2023-transferable,
    title = "Transferable and Efficient: Unifying Dynamic Multi-Domain Product Categorization",
    author = "Gong, Shansan  and
      Zhou, Zelin  and
      Wang, Shuo  and
      Chen, Fengjiao  and
      Song, Xiujie  and
      Cao, Xuezhi  and
      Xian, Yunsen  and
      Zhu, Kenny",
    editor = "Sitaram, Sunayana  and
      Beigman Klebanov, Beata  and
      Williams, Jason D",
    booktitle = "Proceedings of the 61st Annual Meeting of the Association for Computational Linguistics (Volume 5: Industry Track)",
    month = jul,
    year = "2023",
    address = "Toronto, Canada",
    publisher = "Association for Computational Linguistics",
    url = "https://aclanthology.org/2023.acl-industry.46",
    doi = "10.18653/v1/2023.acl-industry.46",
    pages = "476--486",
}

@inproceedings{zou-etal-2024-implicitave,
    title = "{I}mplicit{AVE}: An Open-Source Dataset and Multimodal {LLM}s Benchmark for Implicit Attribute Value Extraction",
    author = "Zou, Henry  and
      Samuel, Vinay  and
      Zhou, Yue  and
      Zhang, Weizhi  and
      Fang, Liancheng  and
      Song, Zihe  and
      Yu, Philip  and
      Caragea, Cornelia",
    editor = "Ku, Lun-Wei  and
      Martins, Andre  and
      Srikumar, Vivek",
    booktitle = "Findings of the Association for Computational Linguistics ACL 2024",
    month = aug,
    year = "2024",
    address = "Bangkok, Thailand and virtual meeting",
    publisher = "Association for Computational Linguistics",
    url = "https://aclanthology.org/2024.findings-acl.20",
    doi = "10.18653/v1/2024.findings-acl.20",
    pages = "338--354",
}

@inproceedings{zhang-etal-2024-discrimination,
    title = "From Discrimination to Generation: Low-Resource Intent Detection with Language Model Instruction Tuning",
    author = "Zhang, Feng  and
      Chen, Wei  and
      Ding, Fei  and
      Gao, Meng  and
      Wang, Tengjiao  and
      Yao, Jiahui  and
      Zheng, Jiabin",
    editor = "Ku, Lun-Wei  and
      Martins, Andre  and
      Srikumar, Vivek",
    booktitle = "Findings of the Association for Computational Linguistics ACL 2024",
    month = aug,
    year = "2024",
    address = "Bangkok, Thailand and virtual meeting",
    publisher = "Association for Computational Linguistics",
    url = "https://aclanthology.org/2024.findings-acl.605",
    doi = "10.18653/v1/2024.findings-acl.605",
    pages = "10167--10183",
}

@inproceedings{ma2024active,
  title={Active generalized category discovery},
  author={Ma, Shijie and Zhu, Fei and Zhong, Zhun and Zhang, Xu-Yao and Liu, Cheng-Lin},
  booktitle={Proceedings of the IEEE/CVF Conference on Computer Vision and Pattern Recognition},
  pages={16890--16900},
  year={2024}
}

@inproceedings{zhang-etal-2023-clusterllm,
    title = "{C}luster{LLM}: Large Language Models as a Guide for Text Clustering",
    author = "Zhang, Yuwei  and
      Wang, Zihan  and
      Shang, Jingbo",
    editor = "Bouamor, Houda  and
      Pino, Juan  and
      Bali, Kalika",
    booktitle = "Proceedings of the 2023 Conference on Empirical Methods in Natural Language Processing",
    month = dec,
    year = "2023",
    address = "Singapore",
    publisher = "Association for Computational Linguistics",
    url = "https://aclanthology.org/2023.emnlp-main.858",
    doi = "10.18653/v1/2023.emnlp-main.858",
    pages = "13903--13920",
}

@inproceedings{xie2016unsupervised,
  title={Unsupervised deep embedding for clustering analysis},
  author={Xie, Junyuan and Girshick, Ross and Farhadi, Ali},
  booktitle={International conference on machine learning},
  pages={478--487},
  year={2016},
  organization={PMLR}
}

@inproceedings{zhong2021neighborhood,
  title={Neighborhood contrastive learning for novel class discovery},
  author={Zhong, Zhun and Fini, Enrico and Roy, Subhankar and Luo, Zhiming and Ricci, Elisa and Sebe, Nicu},
  booktitle={Proceedings of the IEEE/CVF conference on computer vision and pattern recognition},
  pages={10867--10875},
  year={2021}
}

@InProceedings{Zhong_2021_CVPR,
    author    = {Zhong, Zhun and Zhu, Linchao and Luo, Zhiming and Li, Shaozi and Yang, Yi and Sebe, Nicu},
    title     = {OpenMix: Reviving Known Knowledge for Discovering Novel Visual Categories in an Open World},
    booktitle = {Proceedings of the IEEE/CVF Conference on Computer Vision and Pattern Recognition (CVPR)},
    month     = {June},
    year      = {2021},
    pages     = {9462-9470}
}

@inproceedings{wen2023simgcd,
    author    = {Wen, Xin and Zhao, Bingchen and Qi, Xiaojuan},
    title     = {Parametric Classification for Generalized Category Discovery: A Baseline Study},
    booktitle = {Proceedings of the IEEE/CVF International Conference on Computer Vision (ICCV)},
    year      = {2023},
    pages     = {16590-16600}
}

@inproceedings{
    bai2023towards,
    title={Towards Distribution-Agnostic Generalized Category Discovery},
    author={Jianhong Bai and Zuozhu Liu and Hualiang Wang and Ruizhe Chen and Lianrui Mu and Xiaomeng Li and Joey Tianyi Zhou and YANG FENG and Jian Wu and Haoji Hu},
    booktitle={Thirty-seventh Conference on Neural Information Processing Systems},
    year={2023},
    url={https://openreview.net/forum?id=cczH4Xl7Zo}
}

@article{ren2021survey,
  title={A survey of deep active learning},
  author={Ren, Pengzhen and Xiao, Yun and Chang, Xiaojun and Huang, Po-Yao and Li, Zhihui and Gupta, Brij B and Chen, Xiaojiang and Wang, Xin},
  journal={ACM computing surveys (CSUR)},
  volume={54},
  number={9},
  pages={1--40},
  year={2021},
  publisher={ACM New York, NY}
}

@inproceedings{wang2014new,
  title={A new active labeling method for deep learning},
  author={Wang, Dan and Shang, Yi},
  booktitle={2014 International joint conference on neural networks (IJCNN)},
  pages={112--119},
  year={2014},
  organization={IEEE}
}

@article{sener2017active,
  title={Active learning for convolutional neural networks: A core-set approach},
  author={Sener, Ozan and Savarese, Silvio},
  journal={arXiv preprint arXiv:1708.00489},
  year={2017}
}

@inproceedings{
    Ash2020Deep,
    title={Deep Batch Active Learning by Diverse, Uncertain Gradient Lower Bounds},
    author={Jordan T. Ash and Chicheng Zhang and Akshay Krishnamurthy and John Langford and Alekh Agarwal},
    booktitle={International Conference on Learning Representations},
    year={2020},
    url={https://openreview.net/forum?id=ryghZJBKPS}
}

@inproceedings{zou-etal-2023-decrisismb,
    title = "{D}e{C}risis{MB}: Debiased Semi-Supervised Learning for Crisis Tweet Classification via Memory Bank",
    author = "Zou, Henry  and
      Zhou, Yue  and
      Zhang, Weizhi  and
      Caragea, Cornelia",
    editor = "Bouamor, Houda  and
      Pino, Juan  and
      Bali, Kalika",
    booktitle = "Findings of the Association for Computational Linguistics: EMNLP 2023",
    month = dec,
    year = "2023",
    address = "Singapore",
    publisher = "Association for Computational Linguistics",
    url = "https://aclanthology.org/2023.findings-emnlp.406/",
    doi = "10.18653/v1/2023.findings-emnlp.406",
    pages = "6104--6115"
}

@inproceedings{agarwal2020contextual,
  title={Contextual diversity for active learning},
  author={Agarwal, Sharat and Arora, Himanshu and Anand, Saket and Arora, Chetan},
  booktitle={Computer Vision--ECCV 2020: 16th European Conference, Glasgow, UK, August 23--28, 2020, Proceedings, Part XVI 16},
  pages={137--153},
  year={2020},
  organization={Springer}
}

@article{huang2010active,
  title={Active learning by querying informative and representative examples},
  author={Huang, Sheng-Jun and Jin, Rong and Zhou, Zhi-Hua},
  journal={Advances in neural information processing systems},
  volume={23},
  year={2010}
}

@inproceedings{cheng-etal-2023-improving,
    title = "Improving Contrastive Learning of Sentence Embeddings from {AI} Feedback",
    author = "Cheng, Qinyuan  and
      Yang, Xiaogui  and
      Sun, Tianxiang  and
      Li, Linyang  and
      Qiu, Xipeng",
    editor = "Rogers, Anna  and
      Boyd-Graber, Jordan  and
      Okazaki, Naoaki",
    booktitle = "Findings of the Association for Computational Linguistics: ACL 2023",
    month = jul,
    year = "2023",
    address = "Toronto, Canada",
    publisher = "Association for Computational Linguistics",
    url = "https://aclanthology.org/2023.findings-acl.707",
    doi = "10.18653/v1/2023.findings-acl.707",
    pages = "11122--11138",
}

@article{Ma2024ActiveGC,
  title={Active Generalized Category Discovery},
  author={Shijie Ma and Fei Zhu and Zhun Zhong and Xu-Yao Zhang and Cheng-Lin Liu},
  journal={2024 IEEE/CVF Conference on Computer Vision and Pattern Recognition (CVPR)},
  year={2024},
  pages={16890-16900},
  url={https://api.semanticscholar.org/CorpusID:268264086}
}

@article{kuhn1955hungarian,
  title={The Hungarian method for the assignment problem},
  author={Kuhn, Harold W},
  journal={Naval research logistics quarterly},
  volume={2},
  number={1-2},
  pages={83--97},
  year={1955},
  publisher={Wiley Online Library}
}

@inproceedings{zou-etal-2024-eiven,
    title = "{EIVEN}: Efficient Implicit Attribute Value Extraction using Multimodal {LLM}",
    author = "Zou, Henry  and
      Yu, Gavin  and
      Fan, Ziwei  and
      Bu, Dan  and
      Liu, Han  and
      Dai, Peng  and
      Jia, Dongmei  and
      Caragea, Cornelia",
    editor = "Yang, Yi  and
      Davani, Aida  and
      Sil, Avi  and
      Kumar, Anoop",
    booktitle = "Proceedings of the 2024 Conference of the North American Chapter of the Association for Computational Linguistics: Human Language Technologies (Volume 6: Industry Track)",
    month = jun,
    year = "2024",
    address = "Mexico City, Mexico",
    publisher = "Association for Computational Linguistics",
    url = "https://aclanthology.org/2024.naacl-industry.40/",
    doi = "10.18653/v1/2024.naacl-industry.40",
    pages = "453--463"
}

@inproceedings{zhang2021discovering,
  title={Discovering new intents with deep aligned clustering},
  author={Zhang, Hanlei and Xu, Hua and Lin, Ting-En and Lyu, Rui},
  booktitle={Proceedings of the AAAI Conference on Artificial Intelligence},
  volume={35},
  pages={14365--14373},
  year={2021}
}

@inproceedings{zhou-etal-2023-probabilistic,
    title = "A Probabilistic Framework for Discovering New Intents",
    author = "Zhou, Yunhua  and
      Quan, Guofeng  and
      Qiu, Xipeng",
    editor = "Rogers, Anna  and
      Boyd-Graber, Jordan  and
      Okazaki, Naoaki",
    booktitle = "Proceedings of the 61st Annual Meeting of the Association for Computational Linguistics (Volume 1: Long Papers)",
    month = jul,
    year = "2023",
    address = "Toronto, Canada",
    publisher = "Association for Computational Linguistics",
    url = "https://aclanthology.org/2023.acl-long.209/",
    doi = "10.18653/v1/2023.acl-long.209",
    pages = "3771--3784"
}

@article{zhang2023clustering,
  title={A clustering framework for unsupervised and semi-supervised new intent discovery},
  author={Zhang, Hanlei and Xu, Hua and Wang, Xin and Long, Fei and Gao, Kai},
  journal={IEEE Transactions on Knowledge and Data Engineering},
  year={2023},
  publisher={IEEE}
}

@inproceedings{liang-liao-2023-clusterprompt,
    title = "{C}luster{P}rompt: Cluster Semantic Enhanced Prompt Learning for New Intent Discovery",
    author = "Liang, Jinggui  and
      Liao, Lizi",
    editor = "Bouamor, Houda  and
      Pino, Juan  and
      Bali, Kalika",
    booktitle = "Findings of the Association for Computational Linguistics: EMNLP 2023",
    month = dec,
    year = "2023",
    address = "Singapore",
    publisher = "Association for Computational Linguistics",
    url = "https://aclanthology.org/2023.findings-emnlp.702/",
    doi = "10.18653/v1/2023.findings-emnlp.702",
    pages = "10468--10481"
}

@article{Raedt2023IDASID,
  title={IDAS: Intent Discovery with Abstractive Summarization},
  author={Maarten De Raedt and Fr{\'e}deric Godin and Thomas Demeester and Chris Develder},
  journal={ArXiv},
  year={2023},
  volume={abs/2305.19783},
  url={https://api.semanticscholar.org/CorpusID:258987814}
}

@inproceedings{sung-etal-2023-pre,
    title = "Pre-training Intent-Aware Encoders for Zero- and Few-Shot Intent Classification",
    author = "Sung, Mujeen  and
      Gung, James  and
      Mansimov, Elman  and
      Pappas, Nikolaos  and
      Shu, Raphael  and
      Romeo, Salvatore  and
      Zhang, Yi  and
      Castelli, Vittorio",
    editor = "Bouamor, Houda  and
      Pino, Juan  and
      Bali, Kalika",
    booktitle = "Proceedings of the 2023 Conference on Empirical Methods in Natural Language Processing",
    month = dec,
    year = "2023",
    address = "Singapore",
    publisher = "Association for Computational Linguistics",
    url = "https://aclanthology.org/2023.emnlp-main.646/",
    doi = "10.18653/v1/2023.emnlp-main.646",
    pages = "10433--10442"
}

@inproceedings{liang-etal-2024-synergizing,
    title = "Synergizing Large Language Models and Pre-Trained Smaller Models for Conversational Intent Discovery",
    author = "Liang, Jinggui  and
      Liao, Lizi  and
      Fei, Hao  and
      Jiang, Jing",
    editor = "Ku, Lun-Wei  and
      Martins, Andre  and
      Srikumar, Vivek",
    booktitle = "Findings of the Association for Computational Linguistics: ACL 2024",
    month = aug,
    year = "2024",
    address = "Bangkok, Thailand",
    publisher = "Association for Computational Linguistics",
    url = "https://aclanthology.org/2024.findings-acl.840/",
    doi = "10.18653/v1/2024.findings-acl.840",
    pages = "14133--14147"
}

@inproceedings{Devlin2019BERTPO,
  title={BERT: Pre-training of Deep Bidirectional Transformers for Language Understanding},
  author={Jacob Devlin and Ming-Wei Chang and Kenton Lee and Kristina Toutanova},
  booktitle={North American Chapter of the Association for Computational Linguistics},
  year={2019},
  url={https://api.semanticscholar.org/CorpusID:52967399}
}

@inproceedings{demszky2020goemotions,
  title={GoEmotions: A Dataset of Fine-Grained Emotions},
  author={Demszky, Dorottya and Movshovitz-Attias, Dana and Ko, Jeongwoo and Cowen, Alan and Nemade, Gaurav and Ravi, Sujith},
  booktitle={Proceedings of the 58th Annual Meeting of the Association for Computational Linguistics},
  pages={4040--4054},
  year={2020}
}

@inproceedings{rashkin2019towards,
  title={Towards Empathetic Open-domain Conversation Models: A New Benchmark and Dataset},
  author={Rashkin, Hannah and Smith, Eric Michael and Li, Margaret and Boureau, Y-Lan},
  booktitle={Proceedings of the 57th Annual Meeting of the Association for Computational Linguistics},
  pages={5370--5381},
  year={2019}
}

@inproceedings{deng2021ontoed,
  title={OntoED: Low-resource Event Detection with Ontology Embedding},
  author={Deng, Shumin and Zhang, Ningyu and Li, Luoqiu and Hui, Chen and Huaixiao, Tou and Chen, Mosha and Huang, Fei and Chen, Huajun},
  booktitle={Proceedings of the 59th Annual Meeting of the Association for Computational Linguistics and the 11th International Joint Conference on Natural Language Processing (Volume 1: Long Papers)},
  pages={2828--2839},
  year={2021}
}

@inproceedings{han2018fewrel,
  title={FewRel: A Large-Scale Supervised Few-Shot Relation Classification Dataset with State-of-the-Art Evaluation},
  author={Han, Xu and Zhu, Hao and Yu, Pengfei and Wang, Ziyun and Yao, Yuan and Liu, Zhiyuan and Sun, Maosong},
  booktitle={Proceedings of the 2018 Conference on Empirical Methods in Natural Language Processing},
  pages={4803--4809},
  year={2018}
}

\appendix
\label{sec:appendix}

\clearpage



\section{Datasets}
\label{appendix:dataset}
In this section, we provide descriptions, statistics and setup for all evaluated datasets. \\

\noindent \textbf{Dataset Descriptions \& 
 Statistics } BANKING \cite{casanueva-etal-2020-efficient} is a dataset of online banking queries labeled with 77 fine-grained customer intents, such as card arrival, card payment not recognized, pending cash withdrawal and pending card payment. StackOverflow \cite{xu-etal-2015-short} consists of technical questions covering a wide range of programming languages, frameworks, and software tools, and annotated with labels such as MATLAB, WordPress, Apache and Bash. CLINC \cite{larson-etal-2019-evaluation} is a popular dataset for multi-domain intent detection. It contains 150 labels and covers diverse domains such as travel, utility and work. The statistics of these datasets are summarized in Table \ref{tab:dataset_statistics}.\\

\noindent \textbf{Dataset Setup } Following \cite{liang-etal-2024-actively}, we randomly select a specified ratio \{5\%, 10\%, 25\%, 50\%\} of all categories as known categories, denoted as known category ratio (KCR). For each known category, 10\% of the data is selected to form the labeled dataset $\mathcal{D}_l$, while the remaining samples constitute the unlabeled dataset $\mathcal{D}_u$. Our ablation studies in the main text use a default KCR of 25\%, while those in the appendix use a default KCR of 10\%.

\begin{table}[!tbh]
\centering
\resizebox{\columnwidth}{!}{%
\begin{tabular}{@{}cccc@{}}
\toprule
\textbf{Dataset} & \textbf{Domain} & \textbf{\# Categoreis} & \textbf{\# Samples} \\ \midrule
BANKING & Banking & 77 & 13,083 \\
CLINC & Multi-Domain & 150 & 22,500 \\
StackOverflow & Programming & 20 & 22,000 \\ \bottomrule
\end{tabular}%
}
\caption{Dataset statistics.}
\label{tab:dataset_statistics}
\end{table}

\section{Evaluation Metrics}
\label{appendix:metrics}

We utilize three standard evaluation metrics to evaluate the GCD performance: ACC, ARI, and NMI. 
The accuracy metric (ACC) evaluates how well the predicted cluster assignments align with the ground-truth labels, and is formulated as:
$$
ACC = \frac{\sum_{i=1}^N \mathbbm{1}_{y_i=map(\hat{y}_i)}}{N}
$$
where $\hat{y}_i$ represents the predicted label and $y_i$ denotes the ground-truth label for each sample $x_i$. $map(\cdot)$ is a mapping function that uses the Hungarian algorithm \cite{kuhn1955hungarian} to establish a one-to-one correspondence between predicted labels $\hat{y}_i$ and their ground-truth counterparts $y_i$.

The Adjusted Rand Index (ARI) evaluates clustering quality by examining pairwise relationships between predicted and ground-truth cluster assignments. ARI can be calculated as:
$$
\textstyle ARI = \frac{\sum_{i,j} \binom{n_{ij}}{2} - [\sum_i \binom{u_i}{2} \sum_j \binom{v_j}{2}]/\binom{N}{2}}{\frac{1}{2}[\sum_i \binom{u_i}{2} + \sum_j \binom{v_j}{2}] - [\sum_i \binom{u_i}{2} \sum_j \binom{v_j}{2}]/\binom{N}{2}}
$$
where $u_i = \sum_j n_{i,j}$ and $v_j = \sum_i n_{i,j}$, $N$ represents the total sample count, and $n_{i,j}$ is the number of samples simultaneously belonging to the $i^{th}$ predicted cluster and $j^{th}$ ground-truth cluster.

The Normalized Mutual Information (NMI) metric measures the consistency between predicted and ground-truth clustering results by quantifying their mutual information. NMI is defined as:
$$
NMI(\hat{y}, y) = \frac{2 \cdot I(\hat{y}, y)}{H(\hat{y}) + H(y)}
$$
where $\hat{y}$ and $y$ represent the predicted and ground-truth label sets respectively. The mutual information between these sets is denoted by $I(\hat{y}, y)$, while $H(\cdot)$ represents the entropy function.

\section{Hyperparameters}
\label{appendix:hyperparameters}

A complete list of default hyperparameters on all evaluated datasets is provided in Table \ref{tab:hyperparameter}.

\begin{table}[!tbh]
\centering
\resizebox{\columnwidth}{!}{%
\begin{tabular}{@{}c|ccc@{}}
\toprule
\multicolumn{1}{l|}{} & \multicolumn{1}{c|}{\textbf{BANKING}} & \multicolumn{1}{c|}{\textbf{CLINC}} & \textbf{StackOverflow} \\ \midrule
\# Query Samples $v$ & \multicolumn{3}{c}{500} \\ \midrule
Similar Instance Selection Options $M$ & \multicolumn{3}{c}{5} \\ \midrule
\multicolumn{1}{l|}{\# Representatives for Category Characterization} & \multicolumn{3}{c}{10} \\ \midrule
Pseudo Category Selection Candidate Ratio & \multicolumn{3}{c}{0.5} \\ \midrule
Batch Size & \multicolumn{3}{c}{80} \\ \midrule
Learning Rate & \multicolumn{3}{c}{1e-05} \\ \midrule
\# Training Epochs & \multicolumn{3}{c}{25} \\ \midrule
Query Set Update Epoch Interval & \multicolumn{3}{c}{5} \\ \midrule
Alignment Wieght $\lambda$ & \multicolumn{3}{c}{\{0.05, 0.1\}} \\ \midrule
Degree of Freedom $\alpha$ & \multicolumn{3}{c}{1} \\ \midrule
Temperature $\tau$ & \multicolumn{3}{c}{0.07} \\ \midrule
$k$-Nearest Neighbors $k$ & 50 & 50 & 500 \\ \bottomrule
\end{tabular}%
}
\caption{Complete list of default hyperparameters on BANKING, CLINC, StackOverflow datasets.}
\label{tab:hyperparameter}
\end{table}



\begin{table}[!tbh]
\centering
\resizebox{\columnwidth}{!}{%
\begin{tabular}{lcccccc}
\toprule
\textbf{Component} & \textbf{Others} & \textbf{ALUP} & \textbf{Loop} & \textbf{Ours} \\ \midrule
\multicolumn{5}{c}{\cellcolor{gray!25}\textit{\textbf{Diversity}}} \\
Instance-Instance Feedback \ding{172} & \cellcolor{red!20}\ding{55} & \cellcolor{green!25}\ding{51} & \cellcolor{green!25}\ding{51} & \cellcolor{green!25}\ding{51} \\
Cluster-Instance Feedback \ding{173} & \cellcolor{red!20}\ding{55} & \cellcolor{red!20}\ding{55} &\cellcolor{red!20}\ding{55} & \cellcolor{green!25}\ding{51} \\
Cluster-Level Feedback \ding{174} & \cellcolor{red!20}\ding{55} &\cellcolor{red!20}\ding{55} & \cellcolor{green!25}\ding{51} & \cellcolor{green!25}\ding{51} \\
\textbf{(Learning w. Feedback \ding{174})} &\cellcolor{red!20}\ding{55} &\cellcolor{red!20}\ding{55} &\cellcolor{red!20}\ding{55} & \cellcolor{green!25}\ding{51} \\ \midrule
\multicolumn{5}{c}{\cellcolor{gray!25}\textit{\textbf{Quality}}} \\
\textbf{Feedback Quality Investigation} &\cellcolor{red!20}\ding{55} &\cellcolor{red!20}\ding{55} &\cellcolor{red!20}\ding{55} & \cellcolor{green!25}\ding{51} \\
\textbf{Quality Enhancement} &\cellcolor{red!20}\ding{55} &\cellcolor{red!20}\ding{55} &\cellcolor{red!20}\ding{55} & \cellcolor{green!25}\ding{51} \\
\textbf{Feedback Post-Filtering} &\cellcolor{red!20}\ding{55} &\cellcolor{red!20}\ding{55} &\cellcolor{red!20}\ding{55} & \cellcolor{green!25}\ding{51} \\ \bottomrule
\end{tabular}%
}
\caption{Comparison of closely related work in GCD. \textit{ALUP} \cite{liang-etal-2024-actively} and \textit{Loop} \cite{an-etal-2024-generalized} use only \textit{one} type of LLM feedback for \textit{model learning} and \textit{ignore} \textit{feedback quality}. Other work in GCD does not leverage LLM feedback. We are the first to consider both LLM feedback diversity and quality in GCD.}
\label{tab:comp_closely_related_work}
\end{table}

\section{Additional Studies and Results}

\subsection{Different Variants of Prompts}
To investigate the impact of different prompt components on the performance of \MethodName, we conducted experiments with various prompt variants of Category Characterization, as shown in Table \ref{tab:variants_of_prompts}. Notably, we also add demonstrations of some known category names in the full prompt. Our analysis reveals several key findings: (1) The full \MethodName prompt, which includes demonstrations, category names and descriptions, consistently achieves the best performance across all metrics on both the CLINC and Stackoverflow datasets. (2) Removing demonstrations from the prompt leads to a noticeable decrease in performance, particularly in ACC and ARI metrics. This suggests that providing examples helps the LLM better understand the task and generate more accurate category characterizations. (3) Omitting the category name generation results in the most significant performance drop on the CLINC dataset, indicating that concise category labels are particularly important for this dataset. (4) For the Stackoverflow dataset, removing the category description has a smaller impact compared to removing names or demonstrations, suggesting that category names might be more crucial for this technical domain. These results underscore the importance of carefully designed prompts in leveraging LLMs for generalized category discovery. The combination of demonstrations, category names, and descriptions in our prompts contributes to the overall effectiveness of \MethodName across different datasets.

\begin{table}[!tbh]
\centering
\resizebox{\columnwidth}{!}{%
\begin{tabular}{@{}lcccccc@{}}
\toprule
 & \multicolumn{3}{c}{\textbf{CLINC}} & \multicolumn{3}{c}{\textbf{Stackoverflow}} \\ \midrule
\multicolumn{1}{l|}{Prompt Variants} & ACC & NMI & \multicolumn{1}{c|}{ARI} & ACC & NMI & ARI \\ \midrule
\multicolumn{1}{l|}{\MethodName} & \textbf{87.69} & \textbf{95.02} & \multicolumn{1}{c|}{\textbf{82.27}} & \textbf{82.40} & \textbf{79.67} & 62.81 \\
\multicolumn{1}{l|}{w.o. name} & 85.78 & 93.99 & \multicolumn{1}{c|}{79.65} & 79.50 & 76.66 & 61.97 \\
\multicolumn{1}{l|}{w.o. description} & 86.13 & 94.14 & \multicolumn{1}{c|}{79.68} & 81.00 & 76.88 & \textbf{63.22} \\ 
\multicolumn{1}{l|}{w.o. demonstration} & 86.22 & 94.55 & \multicolumn{1}{c|}{80.82} & 79.40 & 76.63 & 61.76 \\
\bottomrule
\end{tabular}%
}
\caption{Different variants of prompts in LLM Category Characterization. Full prompt with demonstrations, category names, and descriptions achieves the best results.}
\label{tab:variants_of_prompts}
\end{table}

\subsection{Influence of Alignment Weight}
In Table \ref{tab:alignment_weight}, we show the influence of alignment weight $\lambda$ on model performance. We observe that the choice of alignment weight greatly influences the model's performance across different datasets. For the CLINC dataset, we observe a substantial improvement in performance as the alignment weight increases from 0 to 0.05, with ACC rising by 3.6 percentage points (from 89.64\% to 93.24\%) and ARI improving by 5.62 percentage points (from 83.82\% to 89.44\%). Similarly, for the Stackoverflow dataset, we see notable gains, particularly in ARI, which increases by 4.12 percentage points (from 74.20\% to 78.32\%) when the alignment weight is set to 0.05.

Nevertheless, setting the alignment weight as 0.05, 0.1 can generally offer us good performance improvements. Beyond this range, the performance boost tends to decline, with a particularly sharp drop observed when the weight is set to 1. This suggests that while the alignment between instance embeddings and category descriptions is crucial for improving model performance, it needs to be carefully balanced with other learning objectives. Too much emphasis on alignment (e.g., weight of 1) can lead to overfitting to the LLM-generated descriptions, potentially at the expense of other important features learned from the data.

\begin{table}[!tbh]
\centering
\resizebox{\columnwidth}{!}{%
\begin{tabular}{@{}ccccccc@{}}
\toprule
 & \multicolumn{3}{c}{\textbf{CLINC}} & \multicolumn{3}{c}{\textbf{Stackoverflow}} \\ \midrule
\multicolumn{1}{c|}{Alignment Weight} & ACC & NMI & \multicolumn{1}{c|}{ARI} & ACC & NMI & ARI \\ \midrule
\multicolumn{1}{c|}{0} & 89.64 & 95.46 & \multicolumn{1}{c|}{83.82} & 86.90 & 81.59 & 74.20 \\
\multicolumn{1}{c|}{0.001} & 91.24 & 95.89 & \multicolumn{1}{c|}{85.84} & 87.30 & 82.18 & 74.79 \\
\multicolumn{1}{c|}{0.05} & \textbf{93.24} & \textbf{96.90} & \multicolumn{1}{c|}{\textbf{89.44}} & \textbf{89.00} & \textbf{84.43} & \textbf{78.32} \\
\multicolumn{1}{c|}{0.1} & 92.93 & 96.87 & \multicolumn{1}{c|}{89.27} & 88.00 & 83.02 & 76.54 \\
\multicolumn{1}{c|}{0.5} & 91.07 & 96.63 & \multicolumn{1}{c|}{87.43} & 86.60 & 81.45 & 74.00 \\
\multicolumn{1}{c|}{1} & 82.49 & 93.68 & \multicolumn{1}{c|}{76.74} & 87.20 & 82.30 & 74.67 \\ \bottomrule
\end{tabular}%
}
\caption{Influence of alignment weight with 50\% known category ratio. Setting the alignment weight as 0.05, 0.1 can generally offer us good performance improvements.}
\label{tab:alignment_weight}
\end{table}

\subsection{Effect of Candidate Ratio}
We now analyze the effect of candidate ratio in Cluster-Instance LLM Feedback in Table \ref{tab:candidate_ratio}. Candidate ratio refers to the number of candidate categories provided for pseudo category selection over the total number of categories in the dataset. We observe that including more candidates can generally improve the model performance on the BANKING dataset as a small number of candidates may potentially omit the most relevant category. However, the performance gets slightly decreased on the CLINC dataset when increasing the ratio from 0.75 to 1, we hypothesize this is because CLINC contains a much larger number of clusters (150) and incorporating all of their descriptions results in a huge prompt length, leading to decreased LLM pseudo category selection performance as the LLM may be overwhelmed by the large number of options.

\begin{table}[!tbh]
\centering
\resizebox{\columnwidth}{!}{%
\begin{tabular}{@{}ccccccc@{}}
\toprule
 & \multicolumn{3}{c}{\textbf{BANKING}} & \multicolumn{3}{c}{\textbf{CLINC}} \\ \midrule
\multicolumn{1}{c|}{Candidate Ratio} & ACC & NMI & \multicolumn{1}{c|}{ARI} & ACC & NMI & ARI \\ \midrule
\multicolumn{1}{c|}{0.1} & 67.56 & 81.51 & \multicolumn{1}{c|}{55.77} & 84.49 & 94.01 & 79.33 \\
\multicolumn{1}{c|}{0.25} & 67.56 & 82.24 & \multicolumn{1}{c|}{57.33} & 87.87 & 94.98 & 82.51 \\
\multicolumn{1}{c|}{0.5} & 67.60 & 82.40 & \multicolumn{1}{c|}{58.38} & 87.42 & 94.30 & 80.55 \\
\multicolumn{1}{c|}{0.75} & 68.02 & 82.57 & \multicolumn{1}{c|}{58.16} & 87.78 & 94.32 & 81.25 \\
\multicolumn{1}{c|}{1} & 69.29 & 82.49 & \multicolumn{1}{c|}{58.37} & 86.80 & 94.51 & 81.13 \\ \bottomrule
\end{tabular}%
}
\caption{Effect of candidate ratio in Cluster-Instance LLM Feedback. Candidate Ratio: the number of candidate categories provided for pseudo category selection over the total number of categories in the dataset.}
\label{tab:candidate_ratio}
\end{table}

\subsection{Generalizability Results}

To show the generalizability of \MethodName, we add experiments on four datasets with other types of data, and the accuracy results are shown in the table below. \MethodName consistently delivers improvements over the most recent SOTA method, Loop \cite{an-etal-2024-generalized}, showing that our approach also works well on datasets with other types of data and domains. We provide short descriptions of the added datasets below: (1) GoEmotions \cite{demszky2020goemotions} is a dataset of Reddit comments labeled with 27 emotions, such as amusement, fear, and gratitude. (2) Empathetic Dialogues \cite{rashkin2019towards} consists of conversations between a speaker and listener and is labeled with 32 fine-grained emotions. (3) FewEvent \cite{deng2021ontoed} is a few‑shot event extraction dataset annotated with 34 event classes. (4) FewRel \cite{han2018fewrel} is a relation classification dataset consisting of labeled sentence-level relation instances across 64 relation types.

\begin{table}[!h]
\resizebox{\columnwidth}{!}{%
\begin{tabular}{@{}lcccccccc@{}}
\toprule
 & \multicolumn{2}{c}{\textbf{EmpatheticDialogues}} & \multicolumn{2}{c}{\textbf{GoEmotions}} & \multicolumn{2}{c}{\textbf{FewRel}} & \multicolumn{2}{c}{\textbf{FewEvent}} \\ \midrule
\textbf{\# Classes} & \multicolumn{2}{c}{\textbf{32}} & \multicolumn{2}{c}{\textbf{27}} & \multicolumn{2}{c}{\textbf{64}} & \multicolumn{2}{c}{\textbf{34}} \\ \midrule
Methods & ACC & NMI & ACC & NMI & ACC & NMI & ACC & NMI \\ \midrule
Loop (ACL 2024) & 36.40 & 44.85 & 32.77 & 33.73 & 60.94 & 75.41 & 42.80 & 62.36 \\
DeLFGCD (Ours) & \textbf{44.05} & \textbf{49.88} & \textbf{37.57} & \textbf{37.14} & \textbf{65.35} & \textbf{77.07} & \textbf{44.75} & \textbf{65.43} \\ \bottomrule
\end{tabular}%
}
\caption{Results on other types of data and domains.}
\label{tab:other_data}
\end{table}

\subsection{Representative Failure Case}

In this section, we provide a representative failure case below, which highlights common errors such as ambiguous cluster overlaps and mismatched category names: \\

\noindent “Given the following utterances and demonstrations of some known category names, return a category name and a short category description to summarize the common intent of these utterances in the format (Category Name: [category\_name], Description: [description]) without explanation. \\

\noindent Demonstrations of Some Known Category Names:['failed\_transfer', 'card\_acceptance', 'activate\_my\_card', 'pending\_card\_payment', 'pending\_top\_up', 'transfer\_not\_received\_by\_recipient', 'reverted\_card\_payment?'] \\

\noindent Utterance 1: how do i use a card to top up? 
Utterance 2: can i top up my card with other cards? 
Utterance 3: is it possible to top up by card? 
Utterance 4: if i want to topup by card, how do i do it? Utterance 5: am i able to add money into my account using my american express?  \\

\noindent LLM Generated Category Name and Description: (Category Name: top\_up\_by\_card, Description: Inquiries about using a card, particularly American Express, to add funds to an account.) \\

\noindent Query Ground Truth Labels: 

['topping\_up\_by\_card', 

'supported\_cards\_and\_currencies', 

'topping\_up\_by\_card', 

'supported\_cards\_and\_currencies', 

'supported\_cards\_and\_currencies']”

\section{Analyses on Category Characterization}
\label{appendix:category_characterization}

This section provides comprehensive analyses to measure the performance of category characterization, and the effect of different representative data selection strategies. \\

\begin{figure*}[!tbh]
    \centering
    \includegraphics[width=\textwidth]{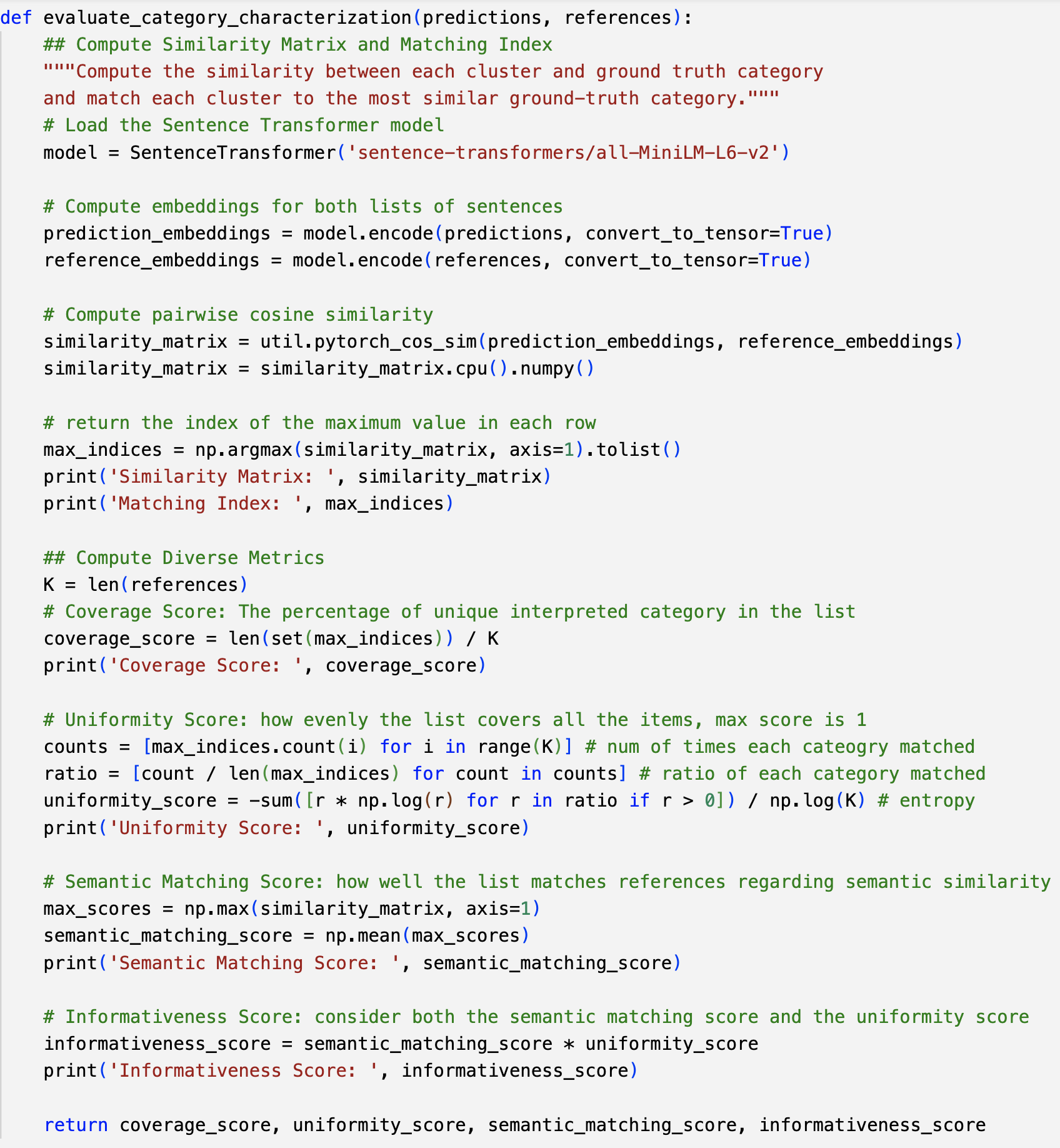} 
    \caption{Implementation of the four evaluation metrics for category characterization: Coverage Score, Uniformity Score, SeMatching Score and Informative Score.}
    \label{fig:cateogy_characterization_evaluation}
\end{figure*}

\noindent \textbf{Evaluation Metrics. } We evaluate the performance of category characterization with the following four metrics, each of which aims to answer different questions or aspects of cluster interpretation: (1) Coverage Score: Do the interpreted/characterized clusters cover all ground-truth categories? (2) Uniformity Score: How evenly do the interpreted clusters cover all ground-truth categories if not covering all ground-truth categories? (3) SeMatching Score: How well does the generated category name \& description match the ground-truth ground-truth categories in terms of semantics similarity?  (4) Informative Score: An overall metric that considers both semantic similarity and uniformity between the interpreted clusters and ground-truth categories. The specific implementation of these metrics is provided in Figure \ref{fig:cateogy_characterization_evaluation}. \\

\noindent \textbf{The effect of different sampling strategies. } We investigate three different sampling strategies for category characterization: (i) \textit{Random}: randomly select $n$ instances from each cluster as representatives; (ii) \textit{Nearest to Center}: For each cluster, select the $n$ instances that are nearest to the K-Means++ cluster center as representatives; (iii) \textit{Sub-KMeans Centroids}: For each cluster, we first perform another K-Means++ clustering and produce $n$ sub-level cluster centroids, which as used as representatives of the original cluster. Table \ref{tab:diff_sampling_strategies} summarizes the evaluation results with these sampling strategies and different numbers of representative samples. Not surprisingly, both \textit{Nearest to Center} and \textit{Sub-KMeans Centroids} sampling strategies perform much better than \textit{Random} sampling. Generally speaking, \textit{Nearest to Center} can achieve slightly better performance than \textit{Sub-KMeans Centroids}. We hypothesize the reason for this is that compared to \textit{Sub-KMeans Centroids}, instances closest to the cluster center are most representative of the cluster and have more coherent semantic meaning, which makes it easier for LLM to produce more accurate and non-overlapping cluster summarization. Furthermore, we observe that including more representatives improves both Coverage and Uniformity Scores, while slightly decreasing the SeMatching Score. This is because having more representatives helps generate unique summarization, but also introduces more noise to make it hard to produce semantically accurate and consistent category descriptions. \\

\noindent \textbf{Performance on different label settings. } Table \ref{tab:diff_cat_label_ratio} summarizes the evaluation results with varying known category ratios and different numbers of labeled data for the \textit{Nearest to Center} sampling strategy. It can be observed that increasing the known category ratio leads to consistently better performance, as more demonstrations of known category names can be provided in the prompt and thus LLM can generate more unique and accurate category names and descriptions. Besides, we can see that as more labeled data is added to known categories, most evaluation scores tend to first increase and then saturate. The SeMatching Score stays roughly the same, indicating that a few number of instances nearest to cluster centers is sufficient to produce semantically similar category names and descriptions with the ground-truth ones.\\

\noindent \textbf{Performance with different LLMs. } We now analyze the impact and costs of using different LLMs, or more specifically, different GPT models.  Table \ref{tab:diff_llms} demonstrates the results. Interestingly, more expensive models, such as \textit{gpt-4} and \textit{gpt-4-turbo}, do necessarily not perform better than cheaper models, such as \textit{gpt-3.5-turbo} and \textit{gpt-4o-mini}, in category characterization. Besides, we can observe that \textit{gpt-4o-mini} performs worse than \textit{gpt-4o} with 1, 10 representatives in Coverage, Uniformity and Informative Scores, but achieves better SeMatching Score and on-par overall performance with 100 representatives. Furthermore, while being 100\texttt{\string~}200 times cheaper than \textit{gpt-4}, \textit{gpt-4o-mini} achieves on-par or better performances in most evaluated scores and settings. The competitive performance and extremely low price of \textit{gpt-4o-mini} render it a good choice to be considered for category characterization. \\

\begin{table*}[!tbh]
\centering
\resizebox{\textwidth}{!}{%
\begin{tabular}{@{}lccccc@{}}
\toprule
\multicolumn{1}{c}{\textbf{Sampling Strategy}} & \textbf{\# Representatives} & \textbf{Coverage} & \textbf{Uniformity} & \textbf{SeMatching} & \textbf{Informative} \\ \midrule
 & 1 & 0.19 & 0.54 & 0.70 & 0.38 \\
 & 3 & \textbf{0.26} & \textbf{0.60} & 0.68 & \textbf{0.41} \\
 & 5 & 0.21 & 0.50 & \textbf{0.71} & 0.35 \\
\multicolumn{1}{c}{\textbf{Random}} & 10 & 0.23 & 0.56 & 0.67 & 0.37 \\
 & 20 & 0.25 & 0.56 & 0.65 & 0.37 \\
 & 50 & 0.23 & 0.54 & 0.65 & 0.35 \\
 & 100 & 0.23 & 0.52 & 0.64 & 0.33 \\ \midrule
 & 1 & 0.47 & 0.71 & \textbf{0.70} & 0.50 \\
 & 3 & 0.62 & 0.81 & 0.69 & 0.56 \\
 & 5 & 0.58 & 0.79 & \textbf{0.70} & 0.55 \\
\multicolumn{1}{c}{\textbf{Nearest to Center}} & 10 & 0.58 & 0.81 & \textbf{0.70} & 0.56 \\
 & 20 & 0.61 & 0.82 & 0.69 & 0.56 \\
 & 50 & 0.57 & 0.82 & 0.68 & 0.56 \\
 & 100 & \textbf{0.65} & \textbf{0.87} & 0.68 & \textbf{0.59} \\ \midrule
 & 1 & 0.45 & 0.71 & \textbf{0.70} & 0.49 \\
 & 3 & 0.55 & 0.79 & 0.66 & 0.52 \\
 & 5 & 0.62 & 0.82 & 0.66 & 0.54 \\
\multicolumn{1}{c}{\textbf{Sub-KMeans Centroids}} & 10 & 0.57 & 0.81 & 0.67 & 0.54 \\
 & 20 & 0.60 & 0.84 & 0.66 & 0.55 \\
 & 50 & 0.60 & 0.84 & 0.66 & 0.55 \\
 & 100 & \textbf{0.65} & \textbf{0.87} & 0.65 & \textbf{0.56} \\ \bottomrule
\end{tabular}%
}
\caption{Effect of different sampling strategies and number of representative samples.}
\label{tab:diff_sampling_strategies}
\end{table*}

\begin{table*}[!tbh]
\centering
\resizebox{\textwidth}{!}{%
\begin{tabular}{@{}lccccc@{}}
\toprule
\multicolumn{1}{c}{\textbf{Known Category Ratio}} & \textbf{Labeled Shot} & \textbf{Coverage} & \textbf{Uniformity} & \textbf{SeMatching} & \textbf{Informative} \\ \midrule
 & 5 & 0.45 & 0.75 & 0.63 & 0.47 \\
 & 10 & 0.58 & 0.80 & \textbf{0.70} & 0.56 \\
\multicolumn{1}{c}{0.1} & 20 & 0.62 & 0.83 & 0.69 & 0.57 \\
 & 50 & \textbf{0.68} & \textbf{0.87} & 0.69 & \textbf{0.60} \\
 & 100 & 0.64 & 0.86 & \textbf{0.67} & 0.58 \\ \midrule
 & 5 & 0.52 & 0.79 & 0.70 & 0.56 \\
 & 10 & 0.66 & 0.87 & 0.69 & 0.60 \\
\multicolumn{1}{c}{0.25} & 20 & \textbf{0.70} & \textbf{0.89} & 0.70 & \textbf{0.62} \\
 & 50 & 0.69 & 0.88 & 0.70 & \textbf{0.62} \\
 & 100 & 0.66 & 0.87 & \textbf{0.71} & 0.61 \\ \midrule
 & 5 & 0.73 & 0.90 & 0.71 & 0.64 \\
 & 10 & 0.77 & 0.92 & 0.71 & 0.66 \\
\multicolumn{1}{c}{0.5} & 20 & 0.79 & 0.92 & 0.71 & 0.66 \\
 & 50 & \textbf{0.81} & \textbf{0.93} & \textbf{0.72} & \textbf{0.67} \\
 & 100 & 0.78 & 0.92 & 0.71 & 0.66 \\ \bottomrule
\end{tabular}%
}
\caption{Performance on different known category ratios and different numbers of labeled data.}
\label{tab:diff_cat_label_ratio}
\end{table*}

\begin{table*}[!tbh]
\centering
\resizebox{\textwidth}{!}{%
\begin{tabular}{@{}lcccccc@{}}
\toprule
\multicolumn{1}{c}{\textbf{\# Representatives}} & \textbf{LLMs} & \textbf{Cost} & \textbf{Coverage} & \textbf{Uniformity} & \textbf{SeMatching} & \textbf{Informative} \\ \midrule
 & gpt-3.5-turbo & \$1/ \$2 & 0.47 & 0.75 & \textbf{0.70} & 0.52 \\
 & gpt-4o-mini & \$0.15 / \$0.6 & 0.48 & 0.72 & \textbf{0.70} & 0.50 \\
\multicolumn{1}{c}{1} & gpt-4o & \$5 / \$15 & \textbf{0.61} & \textbf{0.86} & 0.66 & \textbf{0.57} \\
 & gpt-4-turbo & \$10 / \$30 & 0.45 & 0.74 & 0.65 & 0.48 \\
 & gpt-4 & \$30 / \$60 & 0.45 & 0.71 & \textbf{0.70} & 0.49 \\ \midrule
 & gpt-3.5-turbo & \$1/ \$2 & 0.64 & 0.86 & 0.65 & 0.55 \\
 & gpt-4o-mini & \$0.15 / \$0.6 & 0.58 & 0.80 & \textbf{0.70} & 0.56 \\
\multicolumn{1}{c}{10} & gpt-4o & \$5 / \$15 & \textbf{0.66} & \textbf{0.88} & 0.67 & \textbf{0.59} \\
 & gpt-4-turbo & \$10 / \$30 & 0.62 & 0.86 & 0.64 & 0.55 \\
 & gpt-4 & \$30 / \$60 & 0.65 & 0.86 & 0.65 & 0.56 \\ \midrule
 & gpt-3.5-turbo & \$1/ \$2 & 0.58 & 0.84 & 0.64 & 0.53 \\
 & gpt-4o-mini & \$0.15 / \$0.6 & \textbf{0.66} & \textbf{0.87} & \textbf{0.67} & \textbf{0.58} \\
\multicolumn{1}{c}{100} & gpt-4o & \$5 / \$15 & 0.61 & 0.86 & 0.66 & 0.56 \\
 & gpt-4-turbo & \$10 / \$30 & 0.62 & 0.86 & 0.60 & 0.52 \\
 & gpt-4 & \$30 / \$60 & 0.64 & \textbf{0.87} & 0.61 & 0.53 \\ \bottomrule
\end{tabular}%
}
\caption{Performance with different LLMs.}
\label{tab:diff_llms}
\end{table*}

\section{LLM Feedback Prompts \& Examples}
\label{appendix:prompts}

\begin{figure*}[!tbh]
    \centering
    \includegraphics[width=\textwidth]{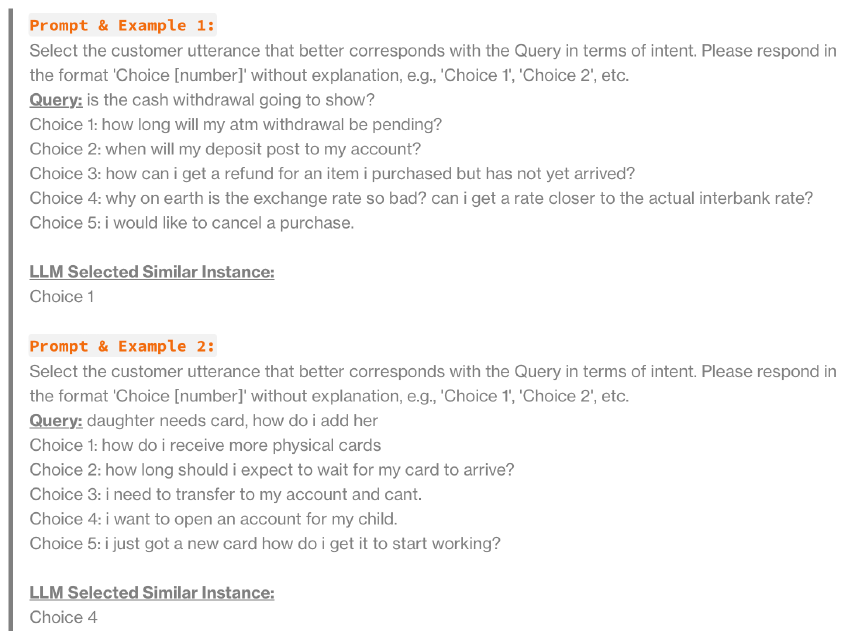} 
    \caption{LLM Feedback 1: Similar Instance Selection Prompt and Example. Click here to return to Section \ref{sec:feedback1}.}
    \label{fig:feedback1_prompt_example}
\end{figure*}

\begin{figure*}[!tbh]
    \centering
    \includegraphics[width=\textwidth]{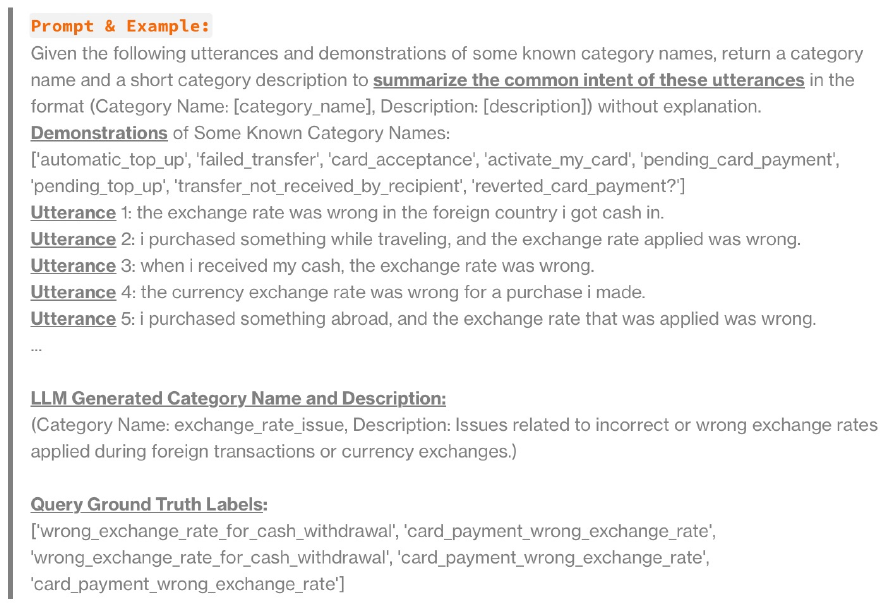} 
    \caption{LLM Feedback 2: Category Characterization Prompt and Example. Click here to return to Section \ref{sec:feedback2}.}
    \label{fig:feedback2_prompt_example}
\end{figure*}

\begin{figure*}[!tbh]
    \centering
    \includegraphics[width=\textwidth]{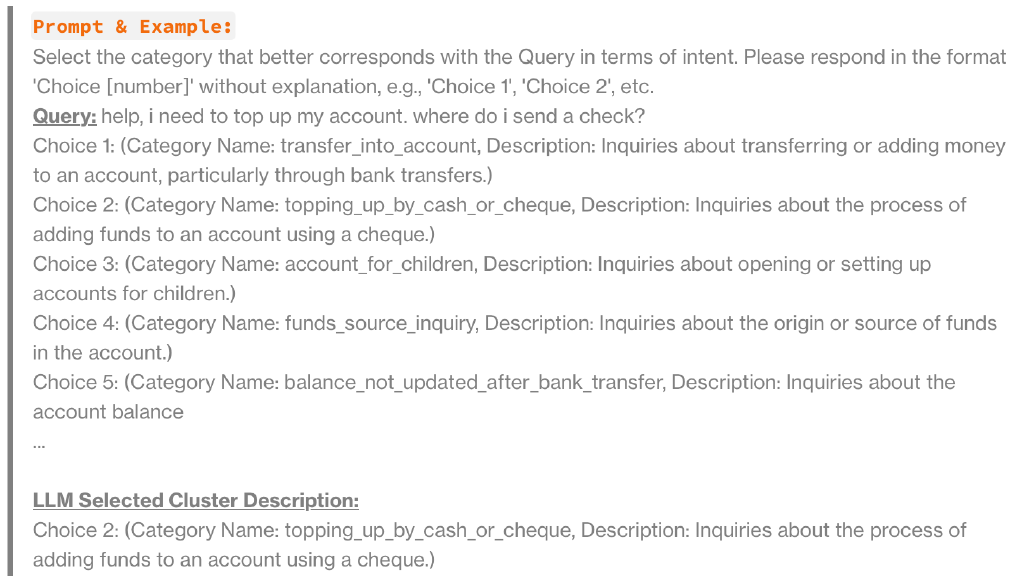} 
    \caption{LLM Feedback 3: Pseudo Category Selection Prompt and Example. Click here to return to Section \ref{sec:feedback3}.}
    \label{fig:feedback3_prompt_example}
\end{figure*}

This section provides all the prompts we used to acquire the three diverse LLM feedback and corresponding examples: (i) Similar Instance Selection - Figure \ref{fig:feedback1_prompt_example}. (ii) Category Characterization - Figure \ref{fig:feedback2_prompt_example}. (iii) Pseudo Category Selection - Figure \ref{fig:feedback3_prompt_example}. Besides, we obtain LLM confidence scores by appending the following sentence in prompts: "Please also show your confidence by providing a probability between 0 and 1.".

\clearpage

\clearpage

\clearpage

\end{document}